\documentclass{article}

\usepackage[preprint]{neurips_data_2024}

\usepackage[utf8]{inputenc} 
\usepackage[T1]{fontenc}    
\usepackage{hyperref}       
\usepackage{url}            
\usepackage{booktabs}       
\usepackage{amsfonts}       
\usepackage{nicefrac}       
\usepackage{xcolor}         
\usepackage{kotex}
\usepackage{booktabs, multirow} 
\usepackage{soul}
\usepackage{graphicx}
\usepackage{wrapfig}
\usepackage{tablefootnote}
\usepackage{anyfontsize}
\usepackage{adjustbox}
\usepackage[most]{tcolorbox}
\usepackage{colortbl}
\usepackage{longtable}

\newcommand{\dashrule}[1][black]{%
  \color{#1}\rule[\dimexpr.5ex-.2pt]{4pt}{.4pt}\xleaders\hbox{\rule{4pt}{0pt}\rule[\dimexpr.5ex-.2pt]{4pt}{.4pt}}\hfill\kern0pt%
}

\title{KMMLU:\\ Measuring Massive Multitask Language Understanding in Korean}

\author{Guijin Son$^{1,2,3}$ \quad Hanwool Lee$^{2,4}$ \quad Sungdong Kim$^{5,6}$ \quad Seungone Kim$^{6,7}$ \\ \textbf{Niklas Muennighoff}$^{8}$ \quad \textbf{Taekyoon Choi}$^{5}$ \quad \textbf{Cheonbok Park}$^{5}$ \\ \textbf{Kang Min Yoo}$^{5}$ \quad \textbf{Stella Biderman}$^{2}$ \\ \\ Yonsei University$^{1}$ \qquad EleutherAI$^{2}$ \qquad OnelineAI$^{3}$ \qquad NCSOFT AI$^{4}$ \\  NAVER Cloud$^{5}$ \qquad KAIST AI$^{6}$ \qquad Carnegie Mellon University$^{7}$ \qquad Contextual AI$^{8}$ \\ \\
\texttt{spthsrbwls123@yonsei.ac.kr} }

\begin{document}

\maketitle

\begin{abstract}
We propose KMMLU, a new Korean benchmark with 35,030 expert-level multiple-choice questions across 45 subjects ranging from humanities to STEM. While prior Korean benchmarks are translated from existing English benchmarks, KMMLU is collected from original Korean exams, capturing linguistic and cultural aspects of the Korean language. We test 27 public and proprietary LLMs and observe the best public model to score 50.5\%, leaving significant room for improvement. This model was primarily trained for English and Chinese, not Korean. Current LLMs tailored to Korean, such as \textsc{Polyglot-Ko}, perform far worse. Surprisingly, even the most capable proprietary LLMs, e.g., \textsc{GPT-4} and \textsc{HyperCLOVA X} do not exceed 60\%. This suggests that further work is needed to improve LLMs for Korean, and we believe KMMLU offers the appropriate tool to track this progress. We make our dataset publicly available on the Hugging Face Hub and integrate the benchmark into EleutherAI's Language Model Evaluation Harness.
\end{abstract}

\section{Introduction}


\begin{figure}[t]
    \centering
    \includegraphics[width=\textwidth]{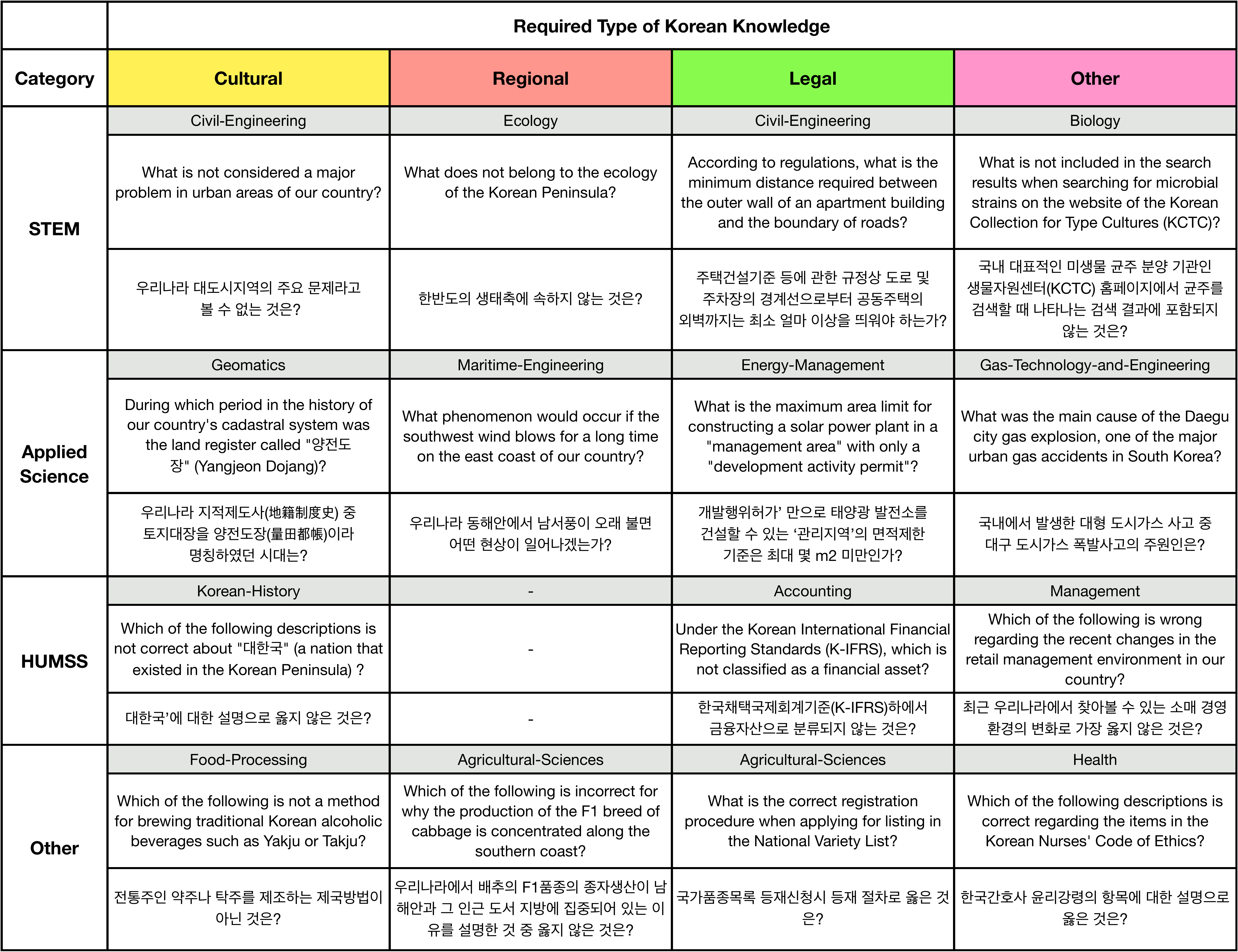}
    \caption{\footnotesize Examples of questions from KMMLU categorized by the type of Korean knowledge required. English translations are added for broader accessibility.}
    \label{fig:examples}
\end{figure}

Recent works often leverage translated versions of MMLU~\citep{hendrycks2020measuring} to evaluate the multilingual capabilities of large language models (LLMs)~\citep{openai2023gpt4, qwen1.5, Chen_MultilingualSIFT_Multilingual_Supervised_2023,zhao2024llama}. However,  naively translating English benchmarks into a language of interest faces critical limitations. First, machine translation can lead to a compromised dataset with issues like unnatural language, typos, and grammatical mistakes~\citep{xia2019generalized,riley2023frmt, yao2023empowering}. Second, MMLU, designed primarily for English speakers, includes content that assumes knowledge of the American legal system or requires familiarity with English slang and culture~\citep{lee2023square,jin2023kobbq,son2023haerae,li2023cmmlu, vmlu}. Thus, while translated versions hint at multilingual proficiency, they often fail to capture the linguistic or cultural aspects that native speakers might consider to be crucial.

To address this issue for the Korean NLP community, we introduce KMMLU, a comprehensive benchmark consisting of 35,030 questions spanning 45 subjects. Unique to KMMLU is its sourcing: \textit{all} questions are derived from Korean exams, ensuring authentic Korean language without any translated material. Additionally, our questions are \textit{localized} to Korea: they reflect the cultural attitudes of Koreans, rather than Westerners (see Figure~\ref{fig:examples}). Our comparative analysis, depicted in Figure~\ref{fig:comp_}, shows that KMMLU surpasses previous translated benchmarks by offering questions that are linguistically natural and steeped in the Korean cultural context. To address the shortcomings of Korean benchmarking, which relies heavily on translated~\citep{park2024open,ham-etal-2020-kornli,jin2023kobbq} or private datasets~\citep{park2024open,park2021klue,lee2024kornat}, we make KMMLU~\footnote{\url{https://huggingface.co/datasets/HAERAE-HUB/KMMLU}} and its corresponding evaluation codes~\footnote{\url{github.com/EleutherAI/lm-evaluation-harness/tree/main/lm_eval/tasks/kmmlu}} publicly available.

We evaluate 27 different LLMs across 5 categories: (1) Multilingual Pretrained Models~\citep{touvron2023llama,yi,bai2023qwen}; (2) Multilingual Chat Models; (3) Korean Pretrained Models~\citep{ko2023polyglot}; (4) Korean Continual Pretrained Models~\citep{l._junbum_2023}; and (5) Proprietary Models including those serviced in Korea~\citep{openai2023gpt4,team2023gemini,kim2021changes}. Our results show significant room for improvement, with \textsc{GPT-4} scoring the highest at 59.95\%. Surprisingly, we see little evidence of a ``curse of multilinguality'' \citep{conneau2019unsupervised,pfeiffer2022lifting} discussed in previous work comparing BLOOM \citep{workshop2022bloom} to monolingual English models \citep{biderman2023pythia,peng2023rwkv}.

Finally, we conduct a detailed analysis to better understand how LLMs utilize Korean knowledge in question-answering. Initially, we observe that, despite \textsc{GPT-4}'s overall excellence, it displays notable gaps in areas demanding \textit{localized knowledge} demonstrating the importance of localizing benchmarks. For example, in Korean History, \textsc{GPT-4}~\citep{openai2023gpt4} achieves a 35\% success rate compared to \textsc{HyperCLOVA X}~\citep{kim2021changes}, a Korean-specific LLM, which scores 44\%.  Notably, \textsc{HyperCLOVA X} is unique in its consistent improvement with the use of Chain-of-Thought (CoT) prompting, indicating the challenge non-Korean LLMs face in producing accurate and reliable Korean explanations.


\section{Related Work}
\subsection{Benchmarks for Large Language Models}

Benchmarks are essential for accurately understanding and tracking the evolving capabilities of LLMs. Traditionally, benchmarks focused on linguistic tasks~\citep{rajpurkar2016squad, wang2019glue, wang2019superglue}, but with the recent surge of more capable LLMs, such approaches have become obsolete. To address this gap, new benchmarks have emerged, focusing on higher-level abilities such as commonsense reasoning~\citep{clark2018think, sakaguchi2021winogrande, zellers2019hellaswag}, mathematical reasoning~\citep{math, cobbe2021training}, code generation~\citep{chen2021evaluating, alpaca_eval}, and multi-turn conversations~\citep{zheng2023judging}. Notably, some efforts have concentrated on evaluating the capabilities via expansive datasets covering a wide range of knowledge-based topics \citep{hendrycks2020measuring,srivastava2022beyond,sawada2023arb}. Most famously, MMLU (Massive Multitask Language Understanding)~\citep{hendrycks2020measuring} spans 57 subjects ranging from basic mathematics to complex areas like law and computer science, evaluating LLMs across various disciplines. While many of these efforts have primarily focused on the English language, there has been progress in adapting and creating similar benchmarks for other languages~\citep{li2023cmmlu, huang2023c, zeng2023measuring, ghahroodi2024khayyam, koto2024arabicmmlu, koto2023large, tam2024improved}.

\subsection{Korean Benchmarks}

\begin{table}[ht]
\centering
\fontsize{8.5}{10}\selectfont
\caption{\footnotesize Overview of Korean benchmarks. The "\# Instances" denots the size of the test set. The \textsc{KLUE} dataset has not released its test set, thus we do not consider it public. The complete release of the \textsc{KorNAT} dataset is scheduled for the future and is currently unavailable as indicated by $\triangle$. Parts of KoBBQ are translated.}
\begin{tabular}{lccccc}
\toprule
\textbf{Korean Benchmarks} & \textbf{Translated} & \textbf{Public?} & \textbf{\# Category} & \textbf{\# Instances} & \textbf{Type} \\
\midrule
\textsc{KorNLI \& KorSTS} & O & O & 2 & 6,389 & Understanding \\
\textsc{KLUE} & X & X & 8 & 38,500 & Understanding \\
\textsc{Ko-H5} & O & X & 5 & 25,700 & Reasoning \\
\textsc{KoBBQ} & $\triangle$ & O & 268 & 76,048 & Bias \\
\textsc{HAE-RAE Bench.} & X & O & 6 & 1,538 & Cultural Knowledge \\
\textsc{CLiCK} & X & O & 11 & 1,995 & Cultural Knowledge \\
\textsc{KorNAT} & X & $\triangle$ & 16 & 10,000 & Cultural Alignment \\
\midrule
KMMLU (Ours) & X & O & 45 & 35,030 & Expert Know. \& Reasoning\\
\bottomrule
\end{tabular}
\label{tab:korbench}
\end{table}


Prior benchmarks for the Korean language specialized on natural language understanding tasks, including natural language inference~\citep{ham-etal-2020-kornli}, machine reading comprehension~\citep{lim2019korquad1}, and hate speech detection~\citep{moon2020beep}. The Korean Language Understanding Evaluation (KLUE) benchmark~\citep{park2021klue}, analogous to the General Language Understanding Evaluation (GLUE)~\citep{wang2019glue}, compiled eight downstream tasks aimed at gauging the comprehension of the Korean language. Nonetheless, these benchmarks exhibited limitations in evaluating reasoning abilities, rendering them insufficient for assessing LLMs. The Ko-H5~\citep{park2024open} aims to overcome these shortcomings by offering an extensive array of reasoning benchmarks, such as HellaSwag~\citep{zellers2019hellaswag}, MMLU~\citep{hendrycks2020measuring}, ARC~\citep{clark2018think}, and TruthfulQA~\citep{lin2021truthfulqa}. Nevertheless, they also rely on machine/human translations that are susceptible to errors. Moreover, their datasets are private and they do not provide evaluation for models larger than 14 billion parameters, thus limiting its transparency and usefulness.

Recent benchmarks developed for Korean have shifted their focus toward preserving the linguistic and cultural nuances, going beyond mere translation of original English benchmarks. In this context, \citet{jin2023kobbq} developed the Korean Bias Benchmark for Question Answering, which is derived from the original BBQ dataset~\citep{parrish-etal-2022-bbq}. This benchmark is specifically adjusted to align with Korean cultural contexts. Additionally, efforts have been made to create authentic Korean datasets from scratch to capture subtle cultural nuances. For instance, the HAE-RAE Benchmark~\citep{son2023haerae} and CLIcK~\citep{kim-etal-2024-click-benchmark} assess cultural and linguistic understanding specific to Korea. Furthermore, KorNAT~\citep{lee2024kornat} was introduced to specifically evaluate alignment with national characteristics of South Korea, focusing on two key aspects: social values and common knowledge.

The KMMLU benchmark extends beyond previous efforts as the first extensive knowledge benchmark covering a broad and deep array of fields, including Biology, Chemistry, Criminal Law, Taxation, Electrical Engineering, Aviation Engineering, and others. Additionally, it is unique in that the dataset, prompts, and evaluation codes are publicly available, ensuring transparent and reproducible assessment, Importantly, it is built entirely from Korean exams, providing a representation of Korean culture and linguistic nuances.

\section{KMMLU}

\begin{wraptable}{r}{0.4\textwidth}
\vspace{-7mm}
\centering
\fontsize{9}{11}\selectfont
\caption{\footnotesize Overview of questions in KMMLU: This table summarizes questions by number of prerequisites for human examinees, whether the question contains negation, and train/validation/test splits.}
\begin{tabular}{lc}
\toprule
\multicolumn{1}{c}{\textbf{Category}} & \textbf{\# Questions} \\
\midrule
\multicolumn{2}{c}{\textit{Prerequisites}} \\
None & 59,909 \\
1 Prerequisite Test & 12,316 \\
2 Prerequisite Tests & 776 \\
2+ Years of Experience & 65,135 \\
4+ Years of Experience & 98,678 \\
9+ Years of Experience & 6,963 \\
\midrule
\multicolumn{2}{c}{\textit{Question Type}} \\
Positive & 207,030 \\
Negation & 36,747 \\
\midrule
\multicolumn{2}{c}{\textit{Split}} \\
Train & 208,522 \\
Validation & 225 \\
Test & 35,030 \\
\midrule
Total & 243,777 \\ 
\bottomrule
\end{tabular}%
\label{tab:kmmlu_stats}
\vspace{-4mm}
\end{wraptable} 

\subsection{Task Overview}

KMMLU is a collection of 35,030 multiple-choice questions spanning 45 categories, including HUMSS (Humanities and Social Science), STEM (science, technology, engineering, and mathematics), Applied Science, and other professional-level knowledge. Within STEM, the focus is on topics emphasizing scientific principles, from the natural and physical sciences to technological and engineering disciplines. Meanwhile, Applied Science encompasses industry-specific subjects such as Aviation Engineering and Maintenance, Gas Technology and Engineering, and Nondestructive Testing. HUMSS covers an extensive range of subjects, including history and psychology, offering in-depth insights into the diverse facets of human society and culture. The remaining subjects that do not fit into any of the three categories are put into Other.

We predominantly source the questions from Korean License Tests, notably, some of the license tests KMMLU draws from exams that requires at least 9 years of industry experience. In addition, KMMLU includes questions that require an understanding of cultural, regional, and legal knowledge to solve, as shown in Figure~\ref{fig:examples}. For further details, see Table~\ref{tab:kmmlu_stats}.

\subsection{Dataset Creation}

Our dataset is a compilation of questions from 533 diverse sources, spanning the Public Service Aptitude Test (PSAT), Korean License Tests, and the College Scholastic Ability Test (CSAT). This collection includes a broad academic spectrum, from high school to professional levels.

Initially, we collected 371,002 questions using automatic crawling. We then implement heuristic filters to erase duplicated samples or parsing errors, including stopwords, regex patterns, and model-based classifiers. Additionally, the format is standardized by excluding questions with fewer than four options and adjusting those with more than four. This filtering reduces the dataset by 34\% to 243,777 questions. The significant reduction in dataset size stems from two main factors: firstly, we prioritize quality over quantity, employing broad filters to eliminate any questionable content, even at the cost of removing some valid samples. Secondly, we observe a high degree of similarity among questions, especially those that are seasonally repeated, resulting in extensive deduplication.

We gather human accuracy data from actual test-takers where available. Approximately 90\% of our dataset's exams include human performance data, with an average accuracy of 62.6\%. Most of the license exams in the dataset require an 80\% score to pass. For the PSAT, the average passing score of the last 5 years has been 83.7\%. Thus, achieving over 80\% on KMMLU can be considered the equivalent of the minimum performance of a human expert, while the best experts are likely to score close to 100\%. The dataset is structured into three components: a training set, a few-shot development set, and a test set. The few-shot development set features five questions per subject to support in-context learning~\citep{brown2020language}. The training set includes 208,522 questions, suitable for both hyperparameter tuning and model training. The test set is a collection of questions with the lowest human accuracies, each subject consisting of 100 instances at a minimum and 35,050 questions in total. However, it is important to note that comparing human accuracy directly may not be desirable due to variations in test origins and test-taker populations, with some groups being more professional than others. While collecting the human data ourselves could potentially address this issue, we could not do so due to budget constraints.

Before finalization, we released the dataset to the public for six months. During this period, we received five issues reported by the community, and 741 instances were modified accordingly. Additionally, the 35,030 questions in the test set underwent manual review to remove copyrighted materials. We replaced 147 instances, including copyrighted materials. No additional errors were identified during this process. Finally, we conduct an analysis based on \citet{xu2024benchmarking}'s method to look for potential data leakages. We observe both open and proprietary LLMs fail to recall the KMMLU benchmark implying low likelihood of benchmark contamination. For details, see Section~\ref{app:cont}.

\subsection{CoT Exemplar Creation}
\label{subsec:cot-exemplar}

\citet{chung2022scaling} devise 5-shot of exemplars to test CoT reasoning over MMLU~\cite{hendrycks2020measuring}~\footnote{\href{https://github.com/jasonwei20/flan-2}{github.com/jasonwei20/flan-2}}. Similarly, we create 5-shot of CoT exemplars for each subject to test models' reasoning capabilities on our benchmark. However, writing an accurate rationale for expert-level tests with various ranges is a difficult problem. Although the ideal solution might be to invite experts for each test, we decide to leverage assistance from various LLMs, considering resource constraints. Specifically, we employ two LLMs, GPT-4 and HyperCLOVA X, with diverse prompt techniques, zero-shot CoT~\cite{kojima2022large} and browsing-augmented CoT~\footnote{It is similar to ReAct prompting~\cite{yao2022react}.}.

First, we elicit rationale and corresponding answers from the LLMs using both prompt techniques. Besides, we utilize a majority voting method, self-consistency~\cite{wang2022self}, over ten reasoning paths obtained by oversampling. As a result, this step produces $4 \times 10$ rationales for each input, i.e., $4 = 2$ LLMs and $2$ prompt types. Then, we choose the top-$4$ rationales ordering by longer and less repetitive output. Finally, authors manually select the most appropriate rationale among the top-$4$ and revise it with thorough inspections if necessary. For quality control, we ensure two workers for each question. We find about 87\% of agreement between two workers at the first iteration. We iteratively validate the remaining conflicted examples. In total, we create $45 \times 5 = 225$ exemplars for the CoT inference within our benchmark. Please see Appendix~\ref{appendix:cot-exemplar-creation} for more details.

\subsection{KMMLU-HARD}

\label{subsec:kmmlu-hard}

KMMLU comprises 35,030 questions, outnumbering its predecessors, MMLU~\citep{hendrycks2020measuring} and CMMLU~\citep{li2023cmmlu}. Thus, in addition to KMMLU, we create KMMLU-HARD for more targeted and efficient evaluation. The KMMLU-HARD subset includes 4,104 questions that at least one of the following models—\textsc{GPT-3.5 Turbo}, \textsc{Gemini Pro}, \textsc{HyperCLOVA X}, and \textsc{GPT-4}—fails to answer correctly. These questions are equally distributed across all categories, each containing 23 to 100 questions.



\section{Experimental Setup}

\subsection{Evaluation Methodology}~\label{sec:4.1}
In our evaluations of LLMs on KMMLU, we employ two distinct settings for a comprehensive comparison. First, the Direct method prompts the model to generate the most plausible option via greedy decoding. In this process, each model generates a response from its entire vocabulary, which makes \(\frac {1}{vocab\_size}\) its random baseline. Second, CoT allows the model to generate text freely and leverages RegEx to parse the results. By generating a sequence of reasoning before the final answer, CoT has succeeded in aiding LLMs to solve reasoning-heavy tasks. Models are set to use greedy decoding for the CoT generation. All evaluations in this paper, regardless of the method, are done in a few-shot setting with five exemplars. Due to hardware constraints, we run our experiments with open models using 8-bit quantization.

\subsection{Models}

In our study, to provide a comprehensive overview of existing LLMs in answering expert-level Korean questions, we evaluate 27 models varying in size, language, and training phase. 

The 27 models include:
\begin{enumerate}
    \item Multilingual Pretrained Models: \textsc{Llama-2} (7B, 13B, 70B)~\citep{touvron2023llama}, \textsc{Qwen} (7B, 14B, 72B)~\citep{bai2023qwen}, and Yi (6B, 34B) \citep{yi};
    \item Multilingual Chat Models: Chat versions of \textsc{Llama-2}, \textsc{Qwen}, and \textsc{Yi};
    \item Korean Pretrained Models: \textsc{Polyglot-Ko} (1.3B, 3.8B, 5.8B, 12.8B)~\citep{ko2023polyglot};
    \item Korean Continual Pretrained Models: \textsc{Llama-2-Ko}~\citep{l._junbum_2023} (7B), and \textsc{Yi-Ko}~\citep{yiko} (6B, 34B);
    \item Proprietary Models: \textsc{GPT-3.5/4}~\citep{openai2023gpt4}\footnote{We use the \textsc{0613} version for both \textsc{GPT} models.}, \textsc{Gemini Pro}~\citep{team2023gemini} and \textsc{HyperCLOVA X}~\cite{kim2021changes}\footnote{We use the \textsc{HCX-L} version.}.
\end{enumerate}    
The inclusion of English \& Chinese bilingual models aims to explore potential spillover effects, given the historical influence of Chinese Hanja on the Korean language. Further details on the models are provided in Appendix~\ref{app:models} and Table~\ref{tab:models}.

\begin{table}[ht]
\centering
\fontsize{9}{11}\selectfont
\caption{\footnotesize Average accuracy(\%) calculated using the Direct method in a 5-shot setting across the entire test set. We report the macro-average accuracy across subjects within each category. The highest-scoring model across the entire table is highlighted in \textbf{bold}, and the best model within each category is \underline{underlined}. Random guessing has an accuracy of 25\% on all subjects. Please see Tables~\ref{tab:ap_ko_p}-\ref{tab:ap_pro} for detailed results.}
\begin{tabular}{lccccc}
\toprule
\multicolumn{1}{c}{\textbf{Model}} & \multicolumn{1}{c}{\textbf{STEM}} & \multicolumn{1}{c}{\textbf{Applied Science}} & \multicolumn{1}{c}{\textbf{HUMSS}} & \multicolumn{1}{c}{\textbf{Other}} & \multicolumn{1}{c}{\textbf{Average}} \\
\midrule
\multicolumn{6}{c}{\textit{\textbf{Multilingual Pretrained Models}}} \\
\midrule
\textsc{LLama-2-7B} & 24.68 & 25.90 & 25.06 & 24.30 & 25.00 \\
\textsc{LLama-2-13B} & 33.81 & 33.86 & 26.26 & 30.86 & 31.26 \\
\textsc{LLama-2-70B} & 41.16 & 38.82 & 41.20 & 40.06 & 40.28 \\
\textsc{Yi-6B} & 35.47 & 34.23 & 33.46 & 35.70 & 34.70 \\
\textsc{Yi-34B} & 44.31 & 40.59 & 47.03 & 43.96 & 43.90 \\
\textsc{Qwen-7B} & 22.74 & 23.83 & 9.44 & 17.59 & 18.52 \\
\textsc{Qwen-14B} & 36.68 & 35.85 & 21.44 & 29.26 & 30.92 \\
\textsc{Qwen-72B} & \underline{50.69} & \underline{47.75} & \underline{54.39} & \underline{50.77} & \underline{50.83} \\
\midrule
\multicolumn{6}{c}{\textit{\textbf{Multilingual Chat Models}}} \\
\midrule
\textsc{LLama-2-7B-Chat} & 28.60 & 29.03 & 26.01 & 27.10 & 27.71 \\
\textsc{LLama-2-13B-Chat} & 30.36 & 29.09 & 26.40 & 29.05 & 28.73 \\
\textsc{LLama-2-70B-Chat} & 35.98 & 34.36 & 32.19 & 35.35 & 34.47 \\
\textsc{Yi-6B-Chat} & 35.58 & 34.55 & 34.39 & 35.95 & 35.11 \\
\textsc{Yi-34B-Chat} & 41.83 & 38.05 & 46.94 & 42.05 & 42.13 \\
\textsc{Qwen-7B-Chat} & 20.26 & 22.16 & 8.67 & 15.70 & 16.82 \\
\textsc{Qwen-14B-Chat} & 32.78 & 33.94 & 19.31 & 26.75 & 28.33 \\
\textsc{Qwen-72B-Chat} & \underline{47.57} & \underline{46.26} & \underline{49.05} & \underline{46.33} & \underline{47.28} \\
\midrule
\multicolumn{6}{c}{\textit{\textbf{Korean Pretrained Models}}} \\
\midrule
\textsc{Polyglot-Ko-1.3B} & 28.77 & 28.02 & 26.99 & 28.11 & 27.97 \\
\textsc{Polyglot-Ko-3.8B} & \underline{29.68} & \underline{31.07} & 26.59 & \underline{29.54} & \underline{29.26} \\
\textsc{Polyglot-Ko-5.8B} & 29.18 & 30.17 & 26.73 & 29.12 & 28.83 \\
\textsc{Polyglot-Ko-12.8B} & 29.27 & 30.08 & \underline{27.08} & 30.55 & \underline{29.26} \\
\midrule
\multicolumn{6}{c}{\textit{\textbf{Korean Continual Pretrained Models}}} \\
\midrule
\textsc{Llama-2-Ko-7B} & 31.60 & 32.50 & 26.33 & 30.00 & 30.10 \\
\textsc{Yi-Ko-6B} & 40.69 & 39.52 & 40.50 & 41.60 & 40.55 \\
\textsc{Yi-Ko-34B} & \underline{50.44} & \underline{46.95} & \underline{53.63} & \underline{51.13} & \underline{50.46} \\
\midrule
\multicolumn{6}{c}{\textit{\textbf{Proprietary Models}}} \\
\midrule
\textsc{GPT-3.5-Turbo} & 44.64 & 42.11 & 40.54 & 42.61 & 42.47 \\
\textsc{Gemini-Pro} & \underline{51.30} & \underline{49.06} & 49.87 & 50.61 & 50.18 \\
\textsc{HyperCLOVA X} & 50.82 & 48.71 & \underline{59.71} & \underline{54.39} & \underline{53.40} \\ 
\textsc{GPT-4} & \textbf{59.95} & \textbf{57.69} & \textbf{63.69} & \textbf{58.65} & \textbf{59.95} \\
\bottomrule
\end{tabular}%
\label{tab:direct}
\end{table}

\section{Evaluation Results}

\begin{wrapfigure}{r}{0.5\textwidth}
    \vspace{-7mm}
    \centering
    \label{fig:budget}
    \includegraphics[width=\linewidth]{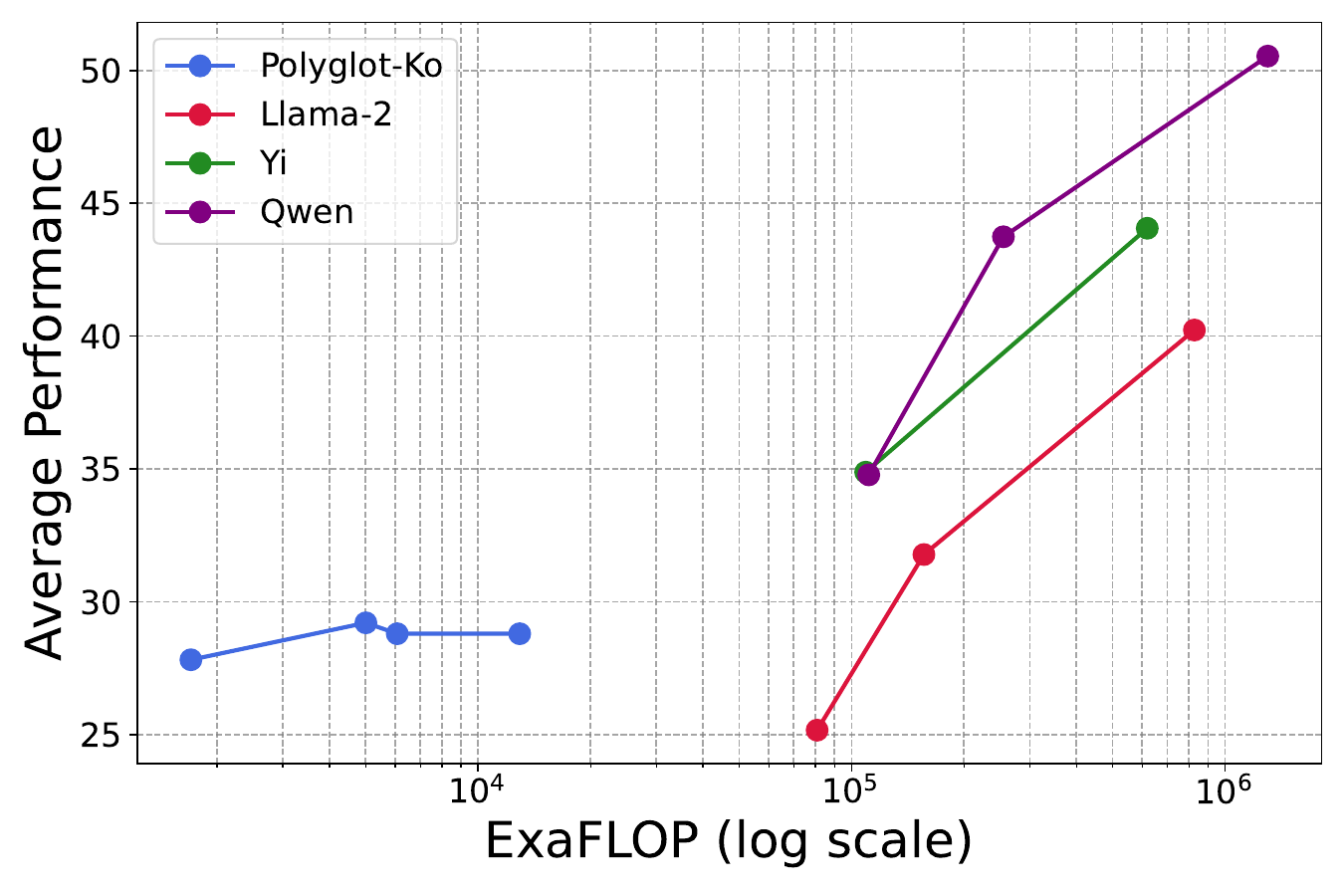}
    \caption{\footnotesize Average performance of \textsc{Polyglot-Ko}, \textsc{Llama-2}, \textsc{Yi}, and \textsc{Qwen} models. ExaFLOP on the x\-axis represents the scale of computational operations, where each unit corresponds to $10^{18}$ floating\-point operations (FLOPs). The total FLOPs are estimated as $6 \times \# param \times \# tokens$~\citep{kaplan2020scaling}.}
\vspace{-4mm}
\end{wrapfigure}

\paragraph{Pretraining Compute}~We compare the performance of 27 LLMs using the Direct method in Table~\ref{tab:direct}. We observe a clear trend across pretrained and fine-tuned models, where those with a larger computing budget exhibit superior performance.\footnote{Unlike other models studied in this paper, the larger Polyglot-Ko models were trained for \textit{fewer} tokens than the smaller ones, explaining the non-monotone performance.} This scaling behavior indicates that increased computing resources - reflected in the number of parameters and the size of the training corpus - enhance a model's capacity to handle complex language tasks more accurately. Notably, despite being trained exclusively in Korean, \textsc{Polyglot-Ko-12.8B}'s performance only marginally exceeds the random baseline of 25\%, is on par with that of the English-centric \textsc{Llama-2-13B}, and lags behind \textsc{Yi} and \textsc{Qwen} models of similar size. This emphasizes the importance of long training runs in achieving high performance: while \textsc{Polyglot-Ko-12.8B} is approximately compute-optimally trained \citep{hoffmann2022training}, the order of magnitude increase in the training data size brings substantial increases in the performance of these non-optimally trained models. This disparity in training resources is further illustrated in figure~\ref{fig:budget}, where Polyglot-Ko's significantly lower training budget compared to its counterparts is evident.


\paragraph{Fine-Tuning}~In Table~\ref{tab:direct}, we also observe that fine-tuning Pretrained Models do not necessarily lead to better performance. In our experiments, models often exhibit minor performance differences between their base and chat versions. This aligns with past studies that suggested fine-tuning methods such as supervised fine-tuning, direct preference optimization, or reinforcement learning to have minor improvements in the knowledge of language models~\citep{bi2024deepseek}. Interestingly, \textsc{Qwen-72B} and \textsc{Llama-2-70B} experience -3.55\% and -5.81\% of performance drop respectively. We suspect that the ability to solve Korean questions in pretrained models of different languages originally stems from their failure to filter out Korean text from their pretraining corpora perfectly. However, datasets used during the post-training process are often curated with greater precision, possibly excluding all non-target languages. Therefore, such might harm the Korean language proficiency of such models.


\paragraph{Multilinguality at Scale}\label{sec:curse} The ``curse of multilinguality''~\citep{conneau2019unsupervised,pfeiffer2022lifting} refers to the apparent decrease in model capabilities when models are trained on multilingual corpora. While the curse can be severe for small models, it has been observed for masked language models that it weakens with scale~\citep{goyal2021larger,pfeiffer2022lifting}. Empirically, this seems not to be the case for the decoder-only \textsc{BLOOM} model~\citep{workshop2022bloom} as several papers have found that monolingual English models substantially outperform BLOOM on English tasks \citep{biderman2023pythia,peng2023rwkv}. In contrast, we find evidence of positive transfer between languages, with large multilingual models like \textsc{Llama-2}, \textsc{Yi}, and \textsc{Qwen} substantially outperforming the monolingual \textsc{Polyglot-Ko}. Though the multilingual models are trained on an order of magnitude more tokens than \textsc{Polyglot-Ko}, they encounter much less Korean text during their pretraining phases. For instance, \textsc{Llama-2} is trained on 2 trillion tokens, with only 0.06\% in Korean, amounting to 1.2 billion tokens. \textsc{Yi} employs a language filter to exclude languages other than Chinese and English, and \textsc{Qwen} mentions that a significant portion of its data is in English and Chinese. In comparison, \textsc{Polyglot-Ko} models are trained on 167 billion to 219 billion tokens depending on the model size. Our results show that scaled decoder-only models acquire capabilities in languages they are severely undertrained in, a finding that aligns with prior work~\citep{muennighoff2023crosslingual}.


\begin{wrapfigure}{r}{0.6\textwidth}
    \vspace{-4mm}
    \centering
    \includegraphics[width=\linewidth]{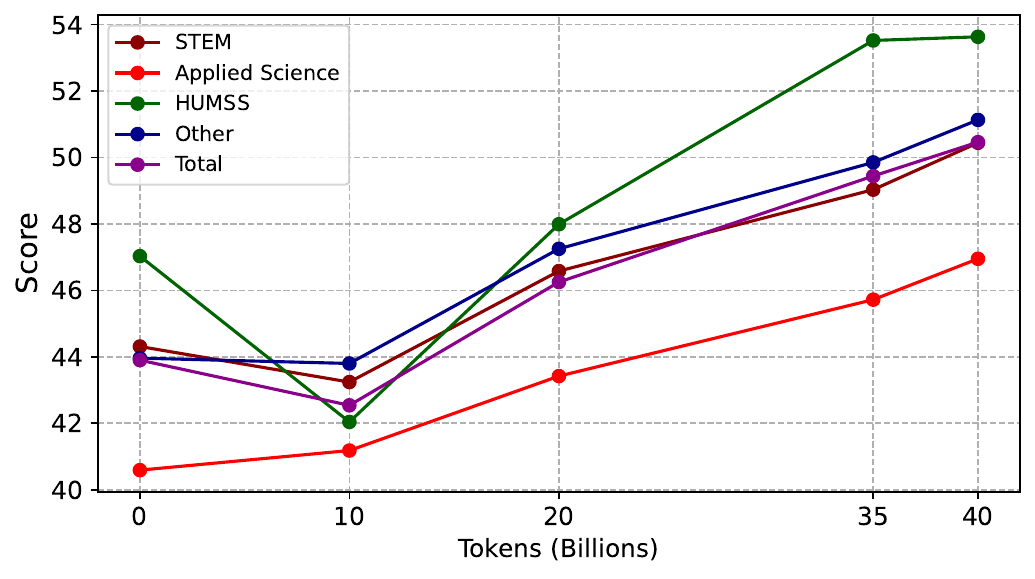}
    \caption{\footnotesize Performance of \textsc{Yi-Ko-34B} on the KMMLU benchmark on each checkpoint. }
    \label{fig:continual}
\vspace{-5mm}
\end{wrapfigure}

\paragraph{Continual Pretraining}~As previously mentioned, models from the \textsc{Yi} series, initially trained for bilingual (English and Chinese) usage, significantly outperform the \textsc{Polyglot-Ko} models. This performance gap further expands with the continual pretraining of these models. In Table~\ref{tab:direct}, we assess the \textsc{Yi-Ko 6B} and \textsc{34B} models, each continually trained for an additional 60 billion and 40 billion tokens, respectively, after expanding their vocabulary to include Korean. Additionally, we analyze the available checkpoints of the \textsc{Yi-Ko-34B} model in Figure~\ref{fig:continual}, noting a consistent performance increase following an initial decline at the first checkpoint. This early drop is likely due to the expanded vocabulary, which can disrupt training initially~\citep{zhao2024llama}. For further details, see Section~\ref{app:contin}.

\section{Analysis}

\subsection{Analysis of Korea-Specific Instances in KMMLU}

\begin{wrapfigure}{l}{0.6\textwidth}
    \vspace{-4mm}
    \centering
    \includegraphics[width=\linewidth]{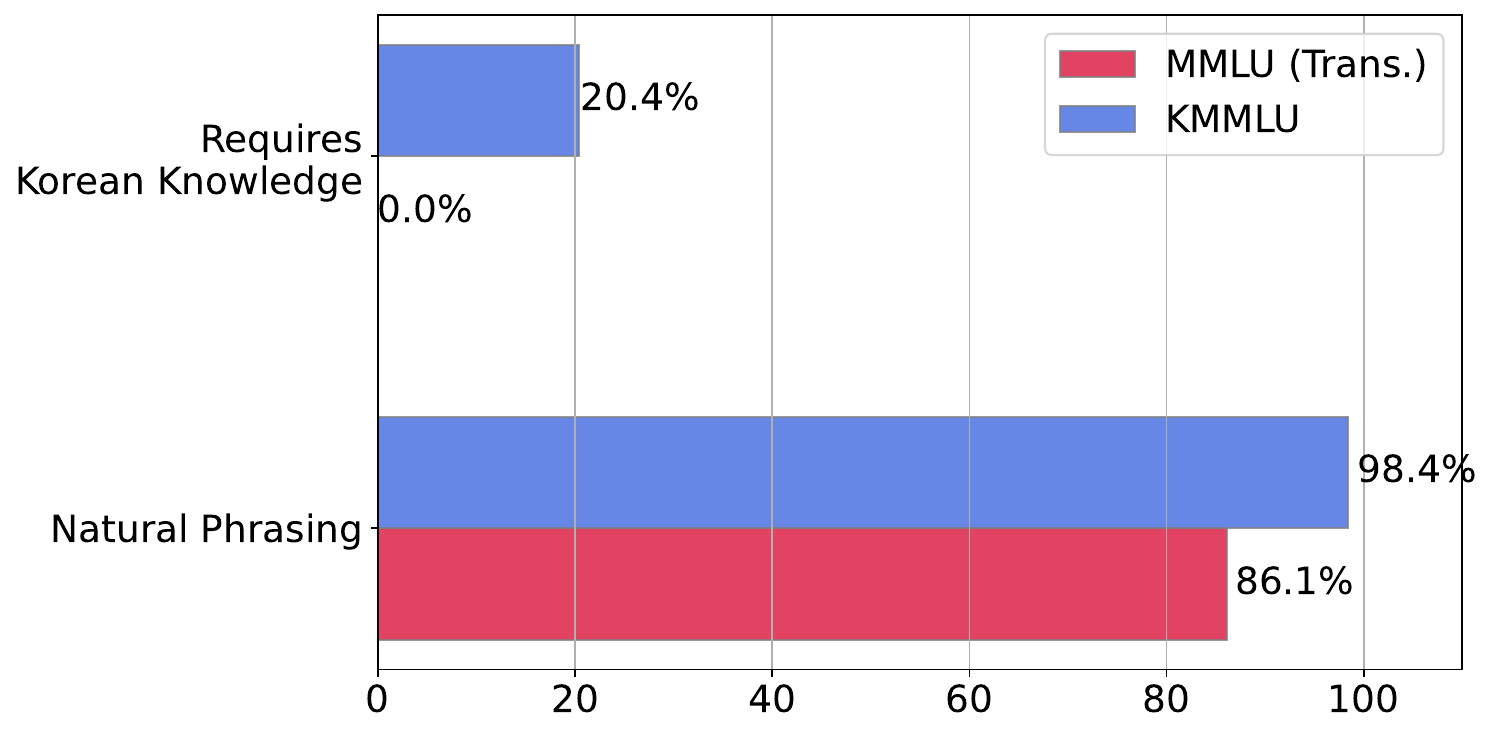}
    \caption{\footnotesize Comparison of MMLU (translated via GPT-4) and KMMLU (ours).}
    \label{fig:comp_}
\vspace{-4mm}
\end{wrapfigure}

To provide a deeper insight into how KMMLU differs from past efforts that translate MMLU~\citep{open-ko-llm-leaderboard,Chen_MultilingualSIFT_Multilingual_Supervised_2023}, we compare the two on two fronts: the naturalness of phrasing and the necessity for specialized Korean knowledge. For the analysis, we randomly selected ten questions from each category within both datasets, resulting in 570 questions from a Korean-translated MMLU and 450 questions from KMMLU—two authors evaluated each question.

Figure~\ref{fig:comp_} reveals a difference in how the two subsets appear to native Korean speakers. KMMLU questions are significantly more natural and culturally relevant, highlighting the limitations of MMLU in reflecting the nuances of the Korean language and cultural specifics. MMLU, derived from American tests, inherently lacks questions about Korean culture. Conversely, 20.4\% of KMMLU requires understanding Korean cultural practices, societal norms, and legal frameworks. This disparity is evident in categories like \textit{``high\_school\_government\_and\_politics''} in MMLU, which lean heavily towards U.S.-centric content, assuming familiarity with the American governmental system, and the \textit{``miscellaneous''} category, which presupposes knowledge of American slang, underscoring the cultural bias embedded within the dataset. However, it is crucial to note that while 20.4\% of KMMLU's content is culturally specific, the remainder broadly assesses a language model's general knowledge, including areas such as mathematics, which are universally applicable and not tied to any particular country's knowledge base.

\subsection{Can Chain-of-Thought prompting improve performance on KMMLU?}

\begin{table}[htb]
\centering
\fontsize{8}{9}\selectfont
\caption{\footnotesize 5-shot accuracy on KMMLU-Hard subset (Section~\ref{subsec:kmmlu-hard}) according to prompting method, Direct and CoT~\cite{wei2022chain}. Please see Table~\ref{tab:ap_cot} for detailed results.}
\begin{tabular}{lcccccccccc}
\toprule

\multicolumn{1}{c}{\textbf{Model}} & \multicolumn{2}{c}{\textbf{STEM}} & \multicolumn{2}{c}{\textbf{Applied Science}} & \multicolumn{2}{c}{\textbf{HUMSS}} & \multicolumn{2}{c}{\textbf{Other}} & \multicolumn{2}{c}{\textbf{Total}} \\
 & Direct & CoT & Direct & CoT & Direct & CoT & Direct & CoT & Direct & CoT \\
 \midrule
Qwen-72B-Chat & 24.36 & 19.00 & 24.25 & 18.67 & 18.52 & 16.50 & 23.09 & 18.38 & 22.59 & 18.18 \\
HyperCLOVA X & 14.36 & 28.00 & 14.58 & 24.83 & 20.62 & \textbf{30.21} & 18.90 & \textbf{25.59} & 17.06 & \textbf{27.11} \\
GPT-3.5-Turbo & 22.36 & 23.27 & 21.00 & 23.67 & 19.74 & 15.35 & 21.30 & 20.25 & 21.10 & 20.70 \\
GPT-4-Turbo & \textbf{28.64} & \textbf{30.91} & \textbf{28.25} & \textbf{34.84} & \textbf{33.37} & 19.68 & \textbf{30.55} & 20.10 & \textbf{30.52} & 25.28 \\
\bottomrule
\end{tabular}%

\label{tab:cot2}
\end{table}

We employ a few-shot CoT prompting~\cite{wei2022chain}, leveraging 5-shot exemplars (Section~\ref{subsec:cot-exemplar}) to examine whether advanced prompting method could improve performance. Since the CoT prompting requires much longer sequence generation than the Direct method, we compare four LLMs based on the KMMLU-Hard subset, considering resource constraints~\footnote{We utilize GPT-4-Turbo \small{(gpt-4-0125-preview)} instead of \textsc{GPT-4} for the same reason.}. In Table~\ref{tab:cot2}, we find that only \textsc{HyperCLOVA X} reliably improves the performances across categories with the CoT prompting, while other LLMs often show degradation with the CoT. In particular, \textsc{GPT-3.5-Turbo} and \textsc{GPT-4-Turbo} show better performances with CoT on STEM and Applied Science, but drastic performance drops on HUMSS. We presume the Korean-specific context in the HUMSS category is relatively hard to generalize by learning other languages, resulting in unfaithful explanations~\cite{turpin2023language}.

\section{Conclusion}

In this work, we introduce the \textbf{KMMLU} Benchmark—a comprehensive compilation of 35,030 expert-level multiple-choice questions spanning 45 subjects, all sourced from original Korean exams without any translated content. Our findings highlight significant room for improvement in the Korean proficiency of state-of-the-art LLMs. We discover that the improvements in the performance of non-Korean LLMs stem from capabilities unrelated to Korean, underscoring the importance of Korean pre-training for better performance in Korea-specific contexts. We expect the KMMLU benchmark to aid researchers in identifying the shortcomings of current models, enabling them to assess and develop better Korean LLMs effectively.

\section{Limitations}~\label{app:limit}
While we put our greatest effort into creating a benchmark with extensive coverage, there are some limitations that future research will need to address. First, due to concerns over copyright issues, we removed a substantial number of questions from the Korean language, medical, and financial domains, thereby creating coverage gaps. Secondly, the recent surge of chat-aligned LLMs has cast doubt on the effectiveness of traditional benchmarks for assessing generative abilities and instruction-following skills. While MMLU continues to be a de facto standard for evaluating a broad range of knowledge, there is a shifting trend towards using dedicated LLM Judges and crowd-sourced human preferences, such as the LMSys Chatbot Arena~\citep{zheng2023judging}, for assessing generative capabilities. Future efforts should aim to expand Korean benchmarking tools to include assessments of generative abilities. Moreover, the potential misuse of benchmarks may pose societal risks. Optimizing solely for benchmarks may create models that perform poorly in real-world applications and should be avoided.




\section*{Acknowledgements}
We thank Zheng Xin Yong and Yukyung Lee for the conversations that helped shape and inform this paper. We also thank Jaehong Lee for his help in evaluating HyperCLOVA X.

\bibliographystyle{natbib}
\bibliography{kmmlu}

\begin{thebibliography}{70}
\expandafter\ifx\csname natexlab\endcsname\relax\def\natexlab#1{#1}\fi

\bibitem[{Bai et~al.(2023)Bai, Bai, Chu, Cui, Dang, Deng, Fan, Ge, Han, Huang et~al.}]{bai2023qwen}
Jinze Bai, Shuai Bai, Yunfei Chu, Zeyu Cui, Kai Dang, Xiaodong Deng, Yang Fan, Wenbin Ge, Yu~Han, Fei Huang, et~al. 2023.
\newblock Qwen technical report.
\newblock \emph{arXiv preprint arXiv:2309.16609}.

\bibitem[{Bi et~al.(2024)Bi, Chen, Chen, Chen, Dai, Deng, Ding, Dong, Du, Fu et~al.}]{bi2024deepseek}
Xiao Bi, Deli Chen, Guanting Chen, Shanhuang Chen, Damai Dai, Chengqi Deng, Honghui Ding, Kai Dong, Qiushi Du, Zhe Fu, et~al. 2024.
\newblock Deepseek llm: Scaling open-source language models with longtermism.
\newblock \emph{arXiv preprint arXiv:2401.02954}.

\bibitem[{Biderman et~al.(2023)Biderman, Schoelkopf, Anthony, Bradley, O’Brien, Hallahan, Khan, Purohit, Prashanth, Raff et~al.}]{biderman2023pythia}
Stella Biderman, Hailey Schoelkopf, Quentin~Gregory Anthony, Herbie Bradley, Kyle O’Brien, Eric Hallahan, Mohammad~Aflah Khan, Shivanshu Purohit, USVSN~Sai Prashanth, Edward Raff, et~al. 2023.
\newblock Pythia: A suite for analyzing large language models across training and scaling.
\newblock In \emph{International Conference on Machine Learning}, pages 2397--2430. PMLR.

\bibitem[{Brown et~al.(2020)Brown, Mann, Ryder, Subbiah, Kaplan, Dhariwal, Neelakantan, Shyam, Sastry, Askell et~al.}]{brown2020language}
Tom Brown, Benjamin Mann, Nick Ryder, Melanie Subbiah, Jared~D Kaplan, Prafulla Dhariwal, Arvind Neelakantan, Pranav Shyam, Girish Sastry, Amanda Askell, et~al. 2020.
\newblock Language models are few-shot learners.
\newblock \emph{Advances in neural information processing systems}, 33:1877--1901.

\bibitem[{Chen et~al.(2021)Chen, Tworek, Jun, Yuan, Pinto, Kaplan, Edwards, Burda, Joseph, Brockman et~al.}]{chen2021evaluating}
Mark Chen, Jerry Tworek, Heewoo Jun, Qiming Yuan, Henrique Ponde de~Oliveira Pinto, Jared Kaplan, Harri Edwards, Yuri Burda, Nicholas Joseph, Greg Brockman, et~al. 2021.
\newblock Evaluating large language models trained on code.
\newblock \emph{arXiv preprint arXiv:2107.03374}.

\bibitem[{Chen et~al.(2023)Chen, Yan, Liang, Jiang, Wu, Yu, Chen, Chen, Zhang, Jianquan, Xiang, and Wang}]{Chen_MultilingualSIFT_Multilingual_Supervised_2023}
Zhihong Chen, Shuo Yan, Juhao Liang, Feng Jiang, Xiangbo Wu, Fei Yu, Guiming~Hardy Chen, Junying Chen, Hongbo Zhang, Li~Jianquan, Wan Xiang, and Benyou Wang. 2023.
\newblock \href {https://github.com/FreedomIntelligence/MultilingualSIFT.git} {{MultilingualSIFT: Multilingual Supervised Instruction Fine-tuning}}.

\bibitem[{Chung et~al.(2022)Chung, Hou, Longpre, Zoph, Tay, Fedus, Li, Wang, Dehghani, Brahma et~al.}]{chung2022scaling}
Hyung~Won Chung, Le~Hou, Shayne Longpre, Barret Zoph, Yi~Tay, William Fedus, Yunxuan Li, Xuezhi Wang, Mostafa Dehghani, Siddhartha Brahma, et~al. 2022.
\newblock Scaling instruction-finetuned language models.
\newblock \emph{arXiv preprint arXiv:2210.11416}.

\bibitem[{Clark et~al.(2018)Clark, Cowhey, Etzioni, Khot, Sabharwal, Schoenick, and Tafjord}]{clark2018think}
Peter Clark, Isaac Cowhey, Oren Etzioni, Tushar Khot, Ashish Sabharwal, Carissa Schoenick, and Oyvind Tafjord. 2018.
\newblock Think you have solved question answering? try arc, the ai2 reasoning challenge.
\newblock \emph{arXiv preprint arXiv:1803.05457}.

\bibitem[{Cobbe et~al.(2021)Cobbe, Kosaraju, Bavarian, Chen, Jun, Kaiser, Plappert, Tworek, Hilton, Nakano et~al.}]{cobbe2021training}
Karl Cobbe, Vineet Kosaraju, Mohammad Bavarian, Mark Chen, Heewoo Jun, Lukasz Kaiser, Matthias Plappert, Jerry Tworek, Jacob Hilton, Reiichiro Nakano, et~al. 2021.
\newblock Training verifiers to solve math word problems.
\newblock \emph{arXiv preprint arXiv:2110.14168}.

\bibitem[{Conneau et~al.(2019)Conneau, Khandelwal, Goyal, Chaudhary, Wenzek, Guzm{\'a}n, Grave, Ott, Zettlemoyer, and Stoyanov}]{conneau2019unsupervised}
Alexis Conneau, Kartikay Khandelwal, Naman Goyal, Vishrav Chaudhary, Guillaume Wenzek, Francisco Guzm{\'a}n, Edouard Grave, Myle Ott, Luke Zettlemoyer, and Veselin Stoyanov. 2019.
\newblock Unsupervised cross-lingual representation learning at scale.
\newblock \emph{arXiv preprint arXiv:1911.02116}.

\bibitem[{Gao et~al.(2023)Gao, Tow, Abbasi, Biderman, Black, DiPofi, Foster, Golding, Hsu, Le~Noac'h, Li, McDonell, Muennighoff, Ociepa, Phang, Reynolds, Schoelkopf, Skowron, Sutawika, Tang, Thite, Wang, Wang, and Zou}]{eval-harness}
Leo Gao, Jonathan Tow, Baber Abbasi, Stella Biderman, Sid Black, Anthony DiPofi, Charles Foster, Laurence Golding, Jeffrey Hsu, Alain Le~Noac'h, Haonan Li, Kyle McDonell, Niklas Muennighoff, Chris Ociepa, Jason Phang, Laria Reynolds, Hailey Schoelkopf, Aviya Skowron, Lintang Sutawika, Eric Tang, Anish Thite, Ben Wang, Kevin Wang, and Andy Zou. 2023.
\newblock \href {https://doi.org/10.5281/zenodo.10256836} {A framework for few-shot language model evaluation}.

\bibitem[{Ghahroodi et~al.(2024)Ghahroodi, Nouri, Sanian, Sahebi, Dastgheib, Asgari, Baghshah, and Rohban}]{ghahroodi2024khayyam}
Omid Ghahroodi, Marzia Nouri, Mohammad~Vali Sanian, Alireza Sahebi, Doratossadat Dastgheib, Ehsaneddin Asgari, Mahdieh~Soleymani Baghshah, and Mohammad~Hossein Rohban. 2024.
\newblock Khayyam challenge (persianmmlu): Is your llm truly wise to the persian language?
\newblock \emph{arXiv preprint arXiv:2404.06644}.

\bibitem[{Goyal et~al.(2021)Goyal, Du, Ott, Anantharaman, and Conneau}]{goyal2021larger}
Naman Goyal, Jingfei Du, Myle Ott, Giri Anantharaman, and Alexis Conneau. 2021.
\newblock Larger-scale transformers for multilingual masked language modeling.
\newblock \emph{arXiv preprint arXiv:2105.00572}.

\bibitem[{Ham et~al.(2020)Ham, Choe, Park, Choi, and Soh}]{ham-etal-2020-kornli}
Jiyeon Ham, Yo~Joong Choe, Kyubyong Park, Ilji Choi, and Hyungjoon Soh. 2020.
\newblock \href {https://doi.org/10.18653/v1/2020.findings-emnlp.39} {{K}or{NLI} and {K}or{STS}: New benchmark datasets for {K}orean natural language understanding}.
\newblock In \emph{Findings of the Association for Computational Linguistics: EMNLP 2020}, pages 422--430, Online. Association for Computational Linguistics.

\bibitem[{Hendrycks et~al.(2020)Hendrycks, Burns, Basart, Zou, Mazeika, Song, and Steinhardt}]{hendrycks2020measuring}
Dan Hendrycks, Collin Burns, Steven Basart, Andy Zou, Mantas Mazeika, Dawn Song, and Jacob Steinhardt. 2020.
\newblock Measuring massive multitask language understanding.
\newblock \emph{arXiv preprint arXiv:2009.03300}.

\bibitem[{Hendrycks et~al.(2021)Hendrycks, Burns, Kadavath, Arora, Basart, Tang, Song, and Steinhardt}]{math}
Dan Hendrycks, Collin Burns, Saurav Kadavath, Akul Arora, Steven Basart, Eric Tang, Dawn Song, and Jacob Steinhardt. 2021.
\newblock Measuring mathematical problem solving with the math dataset.
\newblock \emph{arXiv preprint arXiv:2103.03874}.

\bibitem[{Hoffmann et~al.(2022)Hoffmann, Borgeaud, Mensch, Buchatskaya, Cai, Rutherford, Casas, Hendricks, Welbl, Clark et~al.}]{hoffmann2022training}
Jordan Hoffmann, Sebastian Borgeaud, Arthur Mensch, Elena Buchatskaya, Trevor Cai, Eliza Rutherford, Diego de~Las Casas, Lisa~Anne Hendricks, Johannes Welbl, Aidan Clark, et~al. 2022.
\newblock Training compute-optimal large language models.
\newblock \emph{arXiv preprint arXiv:2203.15556}.

\bibitem[{Hosseini et~al.(2021)Hosseini, Reddy, Bahdanau, Hjelm, Sordoni, and Courville}]{hosseini-etal-2021-understanding}
Arian Hosseini, Siva Reddy, Dzmitry Bahdanau, R~Devon Hjelm, Alessandro Sordoni, and Aaron Courville. 2021.
\newblock \href {https://doi.org/10.18653/v1/2021.naacl-main.102} {Understanding by understanding not: Modeling negation in language models}.
\newblock In \emph{Proceedings of the 2021 Conference of the North American Chapter of the Association for Computational Linguistics: Human Language Technologies}, pages 1301--1312, Online. Association for Computational Linguistics.

\bibitem[{Huang et~al.(2023)Huang, Bai, Zhu, Zhang, Zhang, Su, Liu, Lv, Zhang, Lei et~al.}]{huang2023c}
Yuzhen Huang, Yuzhuo Bai, Zhihao Zhu, Junlei Zhang, Jinghan Zhang, Tangjun Su, Junteng Liu, Chuancheng Lv, Yikai Zhang, Jiayi Lei, et~al. 2023.
\newblock C-eval: A multi-level multi-discipline chinese evaluation suite for foundation models.
\newblock \emph{arXiv preprint arXiv:2305.08322}.

\bibitem[{Jin et~al.(2023)Jin, Kim, Lee, Yoo, Oh, and Lee}]{jin2023kobbq}
Jiho Jin, Jiseon Kim, Nayeon Lee, Haneul Yoo, Alice Oh, and Hwaran Lee. 2023.
\newblock Kobbq: Korean bias benchmark for question answering.
\newblock \emph{arXiv preprint arXiv:2307.16778}.

\bibitem[{Kaplan et~al.(2020)Kaplan, McCandlish, Henighan, Brown, Chess, Child, Gray, Radford, Wu, and Amodei}]{kaplan2020scaling}
Jared Kaplan, Sam McCandlish, Tom Henighan, Tom~B Brown, Benjamin Chess, Rewon Child, Scott Gray, Alec Radford, Jeffrey Wu, and Dario Amodei. 2020.
\newblock Scaling laws for neural language models.
\newblock \emph{arXiv preprint arXiv:2001.08361}.

\bibitem[{Kim et~al.(2024)Kim, Suk, Oh, Yoo, Thorne, and Oh}]{kim-etal-2024-click-benchmark}
Eunsu Kim, Juyoung Suk, Philhoon Oh, Haneul Yoo, James Thorne, and Alice Oh. 2024.
\newblock \href {https://aclanthology.org/2024.lrec-main.296} {{CLI}c{K}: A benchmark dataset of cultural and linguistic intelligence in {K}orean}.
\newblock In \emph{Proceedings of the 2024 Joint International Conference on Computational Linguistics, Language Resources and Evaluation (LREC-COLING 2024)}, pages 3335--3346, Torino, Italia. ELRA and ICCL.

\bibitem[{Ko et~al.(2023)Ko, Yang, Ryu, Choi, Yang, Park et~al.}]{ko2023polyglot}
Hyunwoong Ko, Kichang Yang, Minho Ryu, Taekyoon Choi, Seungmu Yang, Sungho Park, et~al. 2023.
\newblock A technical report for polyglot-ko: Open-source large-scale korean language models.
\newblock \emph{arXiv preprint arXiv:2306.02254}.

\bibitem[{Kojima et~al.(2022)Kojima, Gu, Reid, Matsuo, and Iwasawa}]{kojima2022large}
Takeshi Kojima, Shixiang~Shane Gu, Machel Reid, Yutaka Matsuo, and Yusuke Iwasawa. 2022.
\newblock Large language models are zero-shot reasoners.
\newblock \emph{Advances in neural information processing systems}, 35:22199--22213.

\bibitem[{Koto et~al.(2023)Koto, Aisyah, Li, and Baldwin}]{koto2023large}
Fajri Koto, Nurul Aisyah, Haonan Li, and Timothy Baldwin. 2023.
\newblock Large language models only pass primary school exams in indonesia: A comprehensive test on indommlu.
\newblock \emph{arXiv preprint arXiv:2310.04928}.

\bibitem[{Koto et~al.(2024)Koto, Li, Shatnawi, Doughman, Sadallah, Alraeesi, Almubarak, Alyafeai, Sengupta, Shehata et~al.}]{koto2024arabicmmlu}
Fajri Koto, Haonan Li, Sara Shatnawi, Jad Doughman, Abdelrahman~Boda Sadallah, Aisha Alraeesi, Khalid Almubarak, Zaid Alyafeai, Neha Sengupta, Shady Shehata, et~al. 2024.
\newblock Arabicmmlu: Assessing massive multitask language understanding in arabic.
\newblock \emph{arXiv preprint arXiv:2402.12840}.

\bibitem[{L.~Junbum(2023{\natexlab{a}})}]{yiko}
Taekyoon~Choi L.~Junbum. 2023{\natexlab{a}}.
\newblock \href {https://huggingface.co/beomi/Yi-Ko-6B} {beomi/yi-ko-6b}.

\bibitem[{L.~Junbum(2023{\natexlab{b}})}]{l._junbum_2023}
Taekyoon~Choi L.~Junbum. 2023{\natexlab{b}}.
\newblock \href {https://doi.org/10.57967/hf/1280} {llama-2-koen-13b}.

\bibitem[{Lee et~al.(2023)Lee, Hong, Park, Kim, Cha, Choi, Kim, Kim, Lee, Lim et~al.}]{lee2023square}
Hwaran Lee, Seokhee Hong, Joonsuk Park, Takyoung Kim, Meeyoung Cha, Yejin Choi, Byoung~Pil Kim, Gunhee Kim, Eun-Ju Lee, Yong Lim, et~al. 2023.
\newblock Square: A large-scale dataset of sensitive questions and acceptable responses created through human-machine collaboration.
\newblock \emph{arXiv preprint arXiv:2305.17696}.

\bibitem[{Lee et~al.(2024)Lee, Kim, Kim, Kim, Won, Lee, and Choi}]{lee2024kornat}
Jiyoung Lee, Minwoo Kim, Seungho Kim, Junghwan Kim, Seunghyun Won, Hwaran Lee, and Edward Choi. 2024.
\newblock Kornat: Llm alignment benchmark for korean social values and common knowledge.
\newblock \emph{arXiv preprint arXiv:2402.13605}.

\bibitem[{Li et~al.(2023{\natexlab{a}})Li, Zhang, Koto, Yang, Zhao, Gong, Duan, and Baldwin}]{li2023cmmlu}
Haonan Li, Yixuan Zhang, Fajri Koto, Yifei Yang, Hai Zhao, Yeyun Gong, Nan Duan, and Timothy Baldwin. 2023{\natexlab{a}}.
\newblock \href {http://arxiv.org/abs/2306.09212} {Cmmlu: Measuring massive multitask language understanding in chinese}.

\bibitem[{Li et~al.(2023{\natexlab{b}})Li, Zhang, Dubois, Taori, Gulrajani, Guestrin, Liang, and Hashimoto}]{alpaca_eval}
Xuechen Li, Tianyi Zhang, Yann Dubois, Rohan Taori, Ishaan Gulrajani, Carlos Guestrin, Percy Liang, and Tatsunori~B. Hashimoto. 2023{\natexlab{b}}.
\newblock Alpacaeval: An automatic evaluator of instruction-following models.
\newblock \url{https://github.com/tatsu-lab/alpaca_eval}.

\bibitem[{Lim et~al.(2019)Lim, Kim, and Lee}]{lim2019korquad1}
Seungyoung Lim, Myungji Kim, and Jooyoul Lee. 2019.
\newblock Korquad1. 0: Korean qa dataset for machine reading comprehension.
\newblock \emph{arXiv preprint arXiv:1909.07005}.

\bibitem[{Lin et~al.(2021)Lin, Hilton, and Evans}]{lin2021truthfulqa}
Stephanie Lin, Jacob Hilton, and Owain Evans. 2021.
\newblock Truthfulqa: Measuring how models mimic human falsehoods.
\newblock \emph{arXiv preprint arXiv:2109.07958}.

\bibitem[{Moon et~al.(2020)Moon, Cho, and Lee}]{moon2020beep}
Jihyung Moon, Won~Ik Cho, and Junbum Lee. 2020.
\newblock Beep! korean corpus of online news comments for toxic speech detection.
\newblock \emph{arXiv preprint arXiv:2005.12503}.

\bibitem[{Muennighoff et~al.(2023)Muennighoff, Wang, Sutawika, Roberts, Biderman, Scao, Bari, Shen, Yong, Schoelkopf, Tang, Radev, Aji, Almubarak, Albanie, Alyafeai, Webson, Raff, and Raffel}]{muennighoff2023crosslingual}
Niklas Muennighoff, Thomas Wang, Lintang Sutawika, Adam Roberts, Stella Biderman, Teven~Le Scao, M~Saiful Bari, Sheng Shen, Zheng-Xin Yong, Hailey Schoelkopf, Xiangru Tang, Dragomir Radev, Alham~Fikri Aji, Khalid Almubarak, Samuel Albanie, Zaid Alyafeai, Albert Webson, Edward Raff, and Colin Raffel. 2023.
\newblock \href {http://arxiv.org/abs/2211.01786} {Crosslingual generalization through multitask finetuning}.

\bibitem[{OpenAI(2023)}]{openai2023gpt4}
OpenAI. 2023.
\newblock \href {http://arxiv.org/abs/2303.08774} {Gpt-4 technical report}.

\bibitem[{Park et~al.(2024)Park, Kim, Kim, Cho, Kim, Lee, Kim, and Lee}]{park2024open}
Chanjun Park, Hyeonwoo Kim, Dahyun Kim, Seonghwan Cho, Sanghoon Kim, Sukyung Lee, Yungi Kim, and Hwalsuk Lee. 2024.
\newblock Open ko-llm leaderboard: Evaluating large language models in korean with ko-h5 benchmark.
\newblock \emph{arXiv preprint arXiv:2405.20574}.

\bibitem[{Park et~al.(2023)Park, Lee, Park, Kim, Kim, Cho, Kim, and Lee}]{open-ko-llm-leaderboard}
Chanjun Park, Hwalsuk Lee, Hyunbyung Park, Hyeonwoo Kim, Sanghoon Kim, Seonghwan Cho, Sunghun Kim, and Sukyung Lee. 2023.
\newblock Open ko-llm leaderboard.
\newblock \url{https://huggingface.co/spaces/upstage/open-ko-llm-leaderboard}.

\bibitem[{Park et~al.(2021)Park, Moon, Kim, Cho, Han, Park, Song, Kim, Song, Oh et~al.}]{park2021klue}
Sungjoon Park, Jihyung Moon, Sungdong Kim, Won~Ik Cho, Jiyoon Han, Jangwon Park, Chisung Song, Junseong Kim, Yongsook Song, Taehwan Oh, et~al. 2021.
\newblock Klue: Korean language understanding evaluation.
\newblock \emph{arXiv preprint arXiv:2105.09680}.

\bibitem[{Parrish et~al.(2022)Parrish, Chen, Nangia, Padmakumar, Phang, Thompson, Htut, and Bowman}]{parrish-etal-2022-bbq}
Alicia Parrish, Angelica Chen, Nikita Nangia, Vishakh Padmakumar, Jason Phang, Jana Thompson, Phu~Mon Htut, and Samuel Bowman. 2022.
\newblock \href {https://doi.org/10.18653/v1/2022.findings-acl.165} {{BBQ}: A hand-built bias benchmark for question answering}.
\newblock In \emph{Findings of the Association for Computational Linguistics: ACL 2022}, pages 2086--2105, Dublin, Ireland. Association for Computational Linguistics.

\bibitem[{Peng et~al.(2023)Peng, Alcaide, Anthony, Albalak, Arcadinho, Cao, Cheng, Chung, Grella, GV et~al.}]{peng2023rwkv}
Bo~Peng, Eric Alcaide, Quentin Anthony, Alon Albalak, Samuel Arcadinho, Huanqi Cao, Xin Cheng, Michael Chung, Matteo Grella, Kranthi~Kiran GV, et~al. 2023.
\newblock Rwkv: Reinventing rnns for the transformer era.
\newblock \emph{arXiv preprint arXiv:2305.13048}.

\bibitem[{Pfeiffer et~al.(2022)Pfeiffer, Goyal, Lin, Li, Cross, Riedel, and Artetxe}]{pfeiffer2022lifting}
Jonas Pfeiffer, Naman Goyal, Xi~Victoria Lin, Xian Li, James Cross, Sebastian Riedel, and Mikel Artetxe. 2022.
\newblock Lifting the curse of multilinguality by pre-training modular transformers.
\newblock \emph{arXiv preprint arXiv:2205.06266}.

\bibitem[{Qwen(2024)}]{qwen1.5}
Qwen. 2024.
\newblock \href {https://qwenlm.github.io/blog/qwen1.5/} {Qwen 1.5}.

\bibitem[{Rajpurkar et~al.(2016)Rajpurkar, Zhang, Lopyrev, and Liang}]{rajpurkar2016squad}
Pranav Rajpurkar, Jian Zhang, Konstantin Lopyrev, and Percy Liang. 2016.
\newblock Squad: 100,000+ questions for machine comprehension of text.
\newblock \emph{arXiv preprint arXiv:1606.05250}.

\bibitem[{Riley et~al.(2023)Riley, Dozat, Botha, Garcia, Garrette, Riesa, Firat, and Constant}]{riley2023frmt}
Parker Riley, Timothy Dozat, Jan~A Botha, Xavier Garcia, Dan Garrette, Jason Riesa, Orhan Firat, and Noah Constant. 2023.
\newblock Frmt: A benchmark for few-shot region-aware machine translation.
\newblock \emph{Transactions of the Association for Computational Linguistics}, 11:671--685.

\bibitem[{Sakaguchi et~al.(2021)Sakaguchi, Bras, Bhagavatula, and Choi}]{sakaguchi2021winogrande}
Keisuke Sakaguchi, Ronan~Le Bras, Chandra Bhagavatula, and Yejin Choi. 2021.
\newblock Winogrande: An adversarial winograd schema challenge at scale.
\newblock \emph{Communications of the ACM}, 64(9):99--106.

\bibitem[{Sawada et~al.(2023)Sawada, Paleka, Havrilla, Tadepalli, Vidas, Kranias, Nay, Gupta, and Komatsuzaki}]{sawada2023arb}
Tomohiro Sawada, Daniel Paleka, Alexander Havrilla, Pranav Tadepalli, Paula Vidas, Alexander Kranias, John~J Nay, Kshitij Gupta, and Aran Komatsuzaki. 2023.
\newblock Arb: Advanced reasoning benchmark for large language models.
\newblock \emph{arXiv preprint arXiv:2307.13692}.

\bibitem[{Son et~al.(2023)Son, Lee, Kim, Kim, Lee, Yeom, Jung, Kim, and Kim}]{son2023haerae}
Guijin Son, Hanwool Lee, Suwan Kim, Huiseo Kim, Jaecheol Lee, Je~Won Yeom, Jihyu Jung, Jung~Woo Kim, and Songseong Kim. 2023.
\newblock \href {http://arxiv.org/abs/2309.02706} {Hae-rae bench: Evaluation of korean knowledge in language models}.

\bibitem[{Srivastava et~al.(2022)Srivastava, Rastogi, Rao, Shoeb, Abid, Fisch, Brown, Santoro, Gupta, Garriga-Alonso et~al.}]{srivastava2022beyond}
Aarohi Srivastava, Abhinav Rastogi, Abhishek Rao, Abu Awal~Md Shoeb, Abubakar Abid, Adam Fisch, Adam~R Brown, Adam Santoro, Aditya Gupta, Adri{\`a} Garriga-Alonso, et~al. 2022.
\newblock Beyond the imitation game: Quantifying and extrapolating the capabilities of language models.
\newblock \emph{arXiv preprint arXiv:2206.04615}.

\bibitem[{Tam et~al.(2024)Tam, Pai, Lee, Cheng, and Shuai}]{tam2024improved}
Zhi-Rui Tam, Ya-Ting Pai, Yen-Wei Lee, Sega Cheng, and Hong-Han Shuai. 2024.
\newblock An improved traditional chinese evaluation suite for foundation model.
\newblock \emph{arXiv preprint arXiv:2403.01858}.

\bibitem[{Team et~al.(2023)Team, Anil, Borgeaud, Wu, Alayrac, Yu, Soricut, Schalkwyk, Dai, Hauth et~al.}]{team2023gemini}
Gemini Team, Rohan Anil, Sebastian Borgeaud, Yonghui Wu, Jean-Baptiste Alayrac, Jiahui Yu, Radu Soricut, Johan Schalkwyk, Andrew~M Dai, Anja Hauth, et~al. 2023.
\newblock Gemini: a family of highly capable multimodal models.
\newblock \emph{arXiv preprint arXiv:2312.11805}.

\bibitem[{Touvron et~al.(2023)Touvron, Martin, Stone, Albert, Almahairi, Babaei, Bashlykov, Batra, Bhargava, Bhosale et~al.}]{touvron2023llama}
Hugo Touvron, Louis Martin, Kevin Stone, Peter Albert, Amjad Almahairi, Yasmine Babaei, Nikolay Bashlykov, Soumya Batra, Prajjwal Bhargava, Shruti Bhosale, et~al. 2023.
\newblock Llama 2: Open foundation and fine-tuned chat models.
\newblock \emph{arXiv preprint arXiv:2307.09288}.

\bibitem[{Turpin et~al.(2023)Turpin, Michael, Perez, and Bowman}]{turpin2023language}
Miles Turpin, Julian Michael, Ethan Perez, and Samuel~R Bowman. 2023.
\newblock Language models don't always say what they think: Unfaithful explanations in chain-of-thought prompting.
\newblock \emph{arXiv preprint arXiv:2305.04388}.

\bibitem[{Wang et~al.(2019{\natexlab{a}})Wang, Pruksachatkun, Nangia, Singh, Michael, Hill, Levy, and Bowman}]{wang2019superglue}
Alex Wang, Yada Pruksachatkun, Nikita Nangia, Amanpreet Singh, Julian Michael, Felix Hill, Omer Levy, and Samuel~R. Bowman. 2019{\natexlab{a}}.
\newblock Superglue: A stickier benchmark for general-purpose language understanding systems.
\newblock In \emph{Advances in Neural Information Processing Systems 32: Annual Conference on Neural Information Processing Systems 2019, NeurIPS 2019}.

\bibitem[{Wang et~al.(2019{\natexlab{b}})Wang, Singh, Michael, Hill, Levy, and Bowman}]{wang2019glue}
Alex Wang, Amanpreet Singh, Julian Michael, Felix Hill, Omer Levy, and Samuel~R. Bowman. 2019{\natexlab{b}}.
\newblock \href {http://openreview.net/} {Glue: A multi-task benchmark and analysis platform for natural language understanding}.
\newblock In \emph{7th International Conference on Learning Representations, ICLR 2019}.

\bibitem[{Wang et~al.(2022)Wang, Wei, Schuurmans, Le, Chi, Narang, Chowdhery, and Zhou}]{wang2022self}
Xuezhi Wang, Jason Wei, Dale Schuurmans, Quoc Le, Ed~Chi, Sharan Narang, Aakanksha Chowdhery, and Denny Zhou. 2022.
\newblock Self-consistency improves chain of thought reasoning in language models.
\newblock \emph{arXiv preprint arXiv:2203.11171}.

\bibitem[{Wei et~al.(2022)Wei, Wang, Schuurmans, Bosma, Xia, Chi, Le, Zhou et~al.}]{wei2022chain}
Jason Wei, Xuezhi Wang, Dale Schuurmans, Maarten Bosma, Fei Xia, Ed~Chi, Quoc~V Le, Denny Zhou, et~al. 2022.
\newblock Chain-of-thought prompting elicits reasoning in large language models.
\newblock \emph{Advances in Neural Information Processing Systems}, 35:24824--24837.

\bibitem[{Workshop et~al.(2022)Workshop, Scao, Fan, Akiki, Pavlick, Ili{\'c}, Hesslow, Castagn{\'e}, Luccioni, Yvon et~al.}]{workshop2022bloom}
BigScience Workshop, Teven~Le Scao, Angela Fan, Christopher Akiki, Ellie Pavlick, Suzana Ili{\'c}, Daniel Hesslow, Roman Castagn{\'e}, Alexandra~Sasha Luccioni, Fran{\c{c}}ois Yvon, et~al. 2022.
\newblock Bloom: A 176b-parameter open-access multilingual language model.
\newblock \emph{arXiv preprint arXiv:2211.05100}.

\bibitem[{Xia et~al.(2019)Xia, Kong, Anastasopoulos, and Neubig}]{xia2019generalized}
Mengzhou Xia, Xiang Kong, Antonios Anastasopoulos, and Graham Neubig. 2019.
\newblock Generalized data augmentation for low-resource translation.
\newblock In \emph{Proceedings of the 57th Annual Meeting of the Association for Computational Linguistics}, pages 5786--5796.

\bibitem[{Xu et~al.(2024)Xu, Wang, Fan, and Liu}]{xu2024benchmarking}
Ruijie Xu, Zengzhi Wang, Run-Ze Fan, and Pengfei Liu. 2024.
\newblock Benchmarking benchmark leakage in large language models.
\newblock \emph{arXiv preprint arXiv:2404.18824}.

\bibitem[{Yao et~al.(2023)Yao, Jiang, Yang, and Hu}]{yao2023empowering}
Binwei Yao, Ming Jiang, Diyi Yang, and Junjie Hu. 2023.
\newblock Empowering llm-based machine translation with cultural awareness.
\newblock \emph{arXiv preprint arXiv:2305.14328}.

\bibitem[{Yao et~al.(2022)Yao, Zhao, Yu, Du, Shafran, Narasimhan, and Cao}]{yao2022react}
Shunyu Yao, Jeffrey Zhao, Dian Yu, Nan Du, Izhak Shafran, Karthik Narasimhan, and Yuan Cao. 2022.
\newblock React: Synergizing reasoning and acting in language models.
\newblock \emph{arXiv preprint arXiv:2210.03629}.

\bibitem[{Yoo et~al.(2024)Yoo, Han, In, Jeon, Jeong, Kang, Kim, Kim, Kim, Kim et~al.}]{kim2021changes}
Kang~Min Yoo, Jaegeun Han, Sookyo In, Heewon Jeon, Jisu Jeong, Jaewook Kang, Hyunwook Kim, Kyung-Min Kim, Munhyong Kim, Sungju Kim, et~al. 2024.
\newblock Hyperclova x technical report.
\newblock \emph{arXiv preprint arXiv:2404.01954}.

\bibitem[{Young et~al.(2024)Young, Chen, Li, Huang, Zhang, Zhang, Li, Zhu, Chen, Chang et~al.}]{yi}
Alex Young, Bei Chen, Chao Li, Chengen Huang, Ge~Zhang, Guanwei Zhang, Heng Li, Jiangcheng Zhu, Jianqun Chen, Jing Chang, et~al. 2024.
\newblock Yi: Open foundation models by 01. ai.
\newblock \emph{arXiv preprint arXiv:2403.04652}.

\bibitem[{ZaloAI-JAIST(2023)}]{vmlu}
ZaloAI-JAIST. 2023.
\newblock \href {https://github.com/ZaloAI-Jaist/VMLU/} {Vmlu}.

\bibitem[{Zellers et~al.(2019)Zellers, Holtzman, Bisk, Farhadi, and Choi}]{zellers2019hellaswag}
Rowan Zellers, Ari Holtzman, Yonatan Bisk, Ali Farhadi, and Yejin Choi. 2019.
\newblock Hellaswag: Can a machine really finish your sentence?
\newblock \emph{arXiv preprint arXiv:1905.07830}.

\bibitem[{Zeng(2023)}]{zeng2023measuring}
Hui Zeng. 2023.
\newblock Measuring massive multitask chinese understanding.
\newblock \emph{arXiv preprint arXiv:2304.12986}.

\bibitem[{Zhao et~al.(2024)Zhao, Zhang, Zhang, Gui, and Huang}]{zhao2024llama}
Jun Zhao, Zhihao Zhang, Qi~Zhang, Tao Gui, and Xuanjing Huang. 2024.
\newblock Llama beyond english: An empirical study on language capability transfer.
\newblock \emph{arXiv preprint arXiv:2401.01055}.

\bibitem[{Zheng et~al.(2023)Zheng, Chiang, Sheng, Zhuang, Wu, Zhuang, Lin, Li, Li, Xing et~al.}]{zheng2023judging}
Lianmin Zheng, Wei-Lin Chiang, Ying Sheng, Siyuan Zhuang, Zhanghao Wu, Yonghao Zhuang, Zi~Lin, Zhuohan Li, Dacheng Li, Eric Xing, et~al. 2023.
\newblock Judging llm-as-a-judge with mt-bench and chatbot arena.
\newblock \emph{arXiv preprint arXiv:2306.05685}.

\end{thebibliography}

\appendix











\newpage
\section{Dataset Details}\label{app:dataset}
In this section, we provide additional details on the KMMLU benchmark.

\subsection{Category Distribution}
\begin{figure}[ht]
    \centering
    \includegraphics[width=\linewidth]{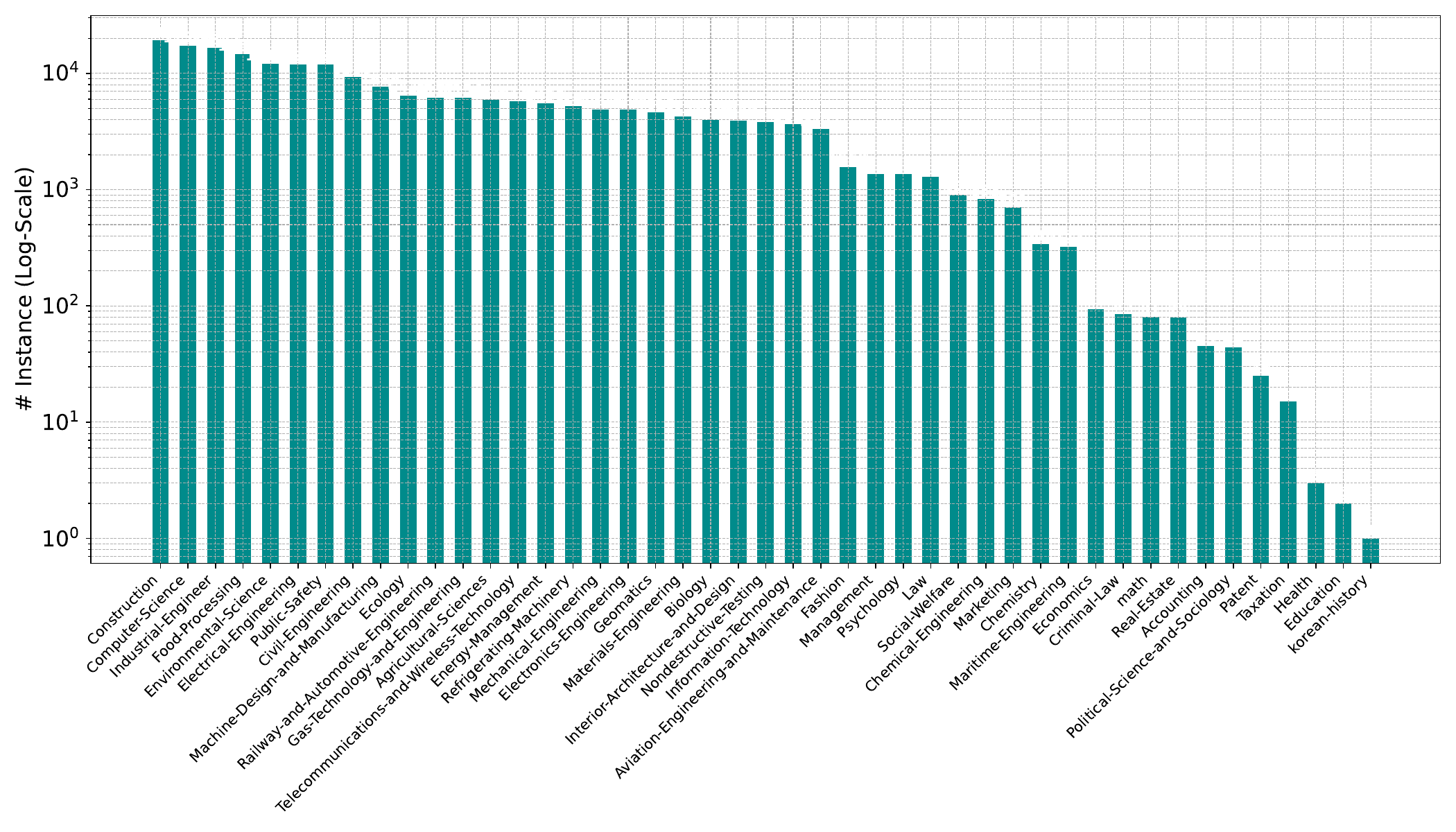}
    \caption{\footnotesize Sub-Category distribution of the train set in log-scale.}
    \label{fig:train_stats}
\end{figure}

\begin{figure}[ht]
    \centering
    \includegraphics[width=\linewidth]{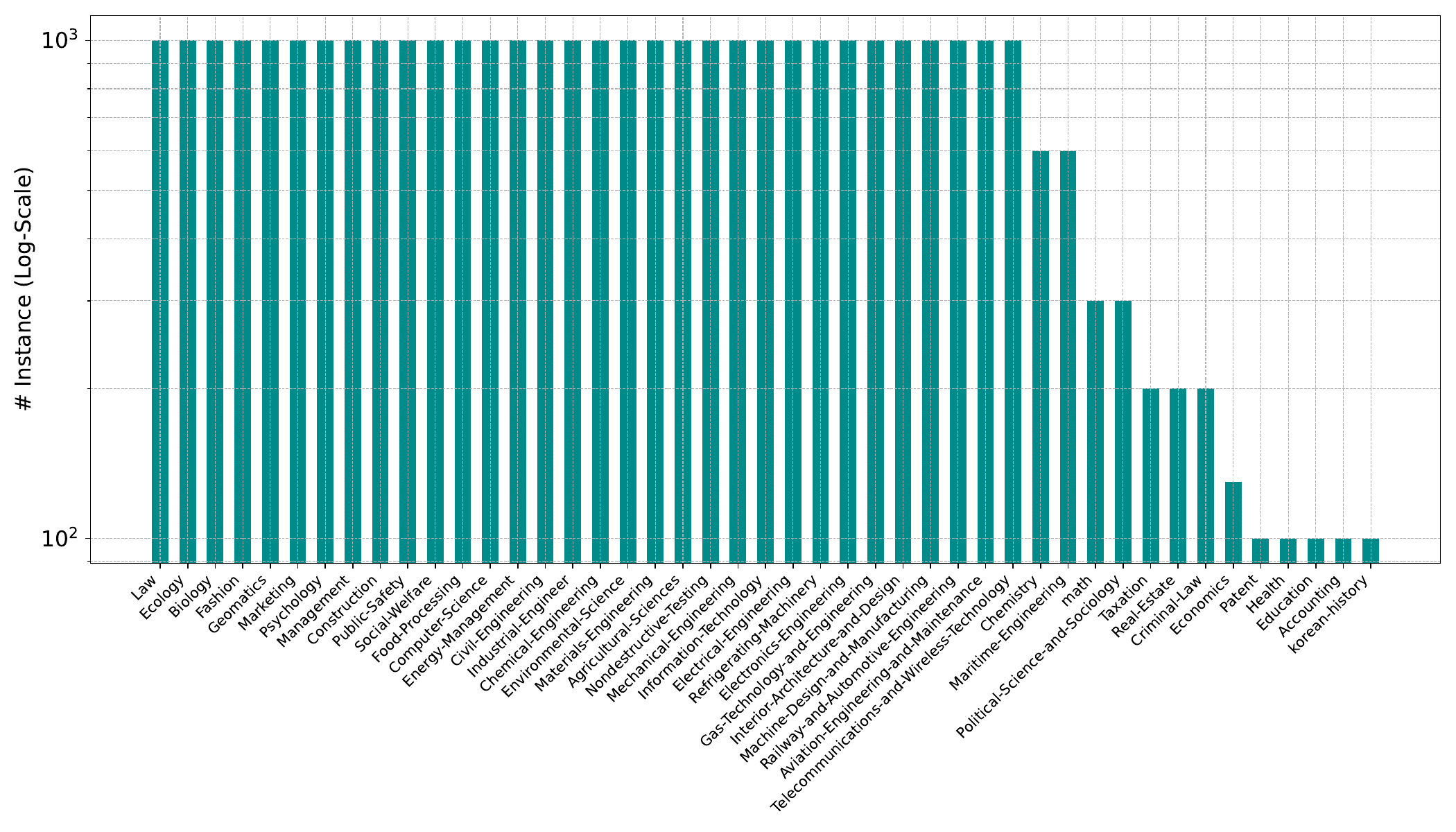}
    \caption{\footnotesize Sub-Category distribution of the test set in log-scale.}
    \label{fig:test_stats}
\end{figure}

The KMMLU dataset consists of three subsets Train, Validation and Test. In Figures~\ref{fig:train_stats} and \ref{fig:test_stats} we visualize the distribution of each sub-category for the train and test set respectively. The validation set contains 5 instances for all sub-categories.

\subsection{Source Distribution}

\begin{wraptable}{r}{0.4\textwidth}
\vspace{-4mm}
\caption{Source distribution}
\centering
\fontsize{9}{11}\selectfont
\begin{tabular}{cc}
\toprule
\textbf{Source} & \textbf{\# Instances} \\ \midrule
License Tests & 235,976 \\
PSAT & 7373 \\
CSAT & 428 \\
\bottomrule
\end{tabular}
\label{tab:source_stats}
\end{wraptable}

KMMLU sources questions from 533 diverse exams, including the PSAT, Korean License Tests, and the CSAT. Table~\ref{tab:source_stats} presents the distribution of the sources, with License Tests comprising the majority at 235,976 samples. 

\begin{figure}[ht]
    \centering
    \includegraphics[width=\linewidth]{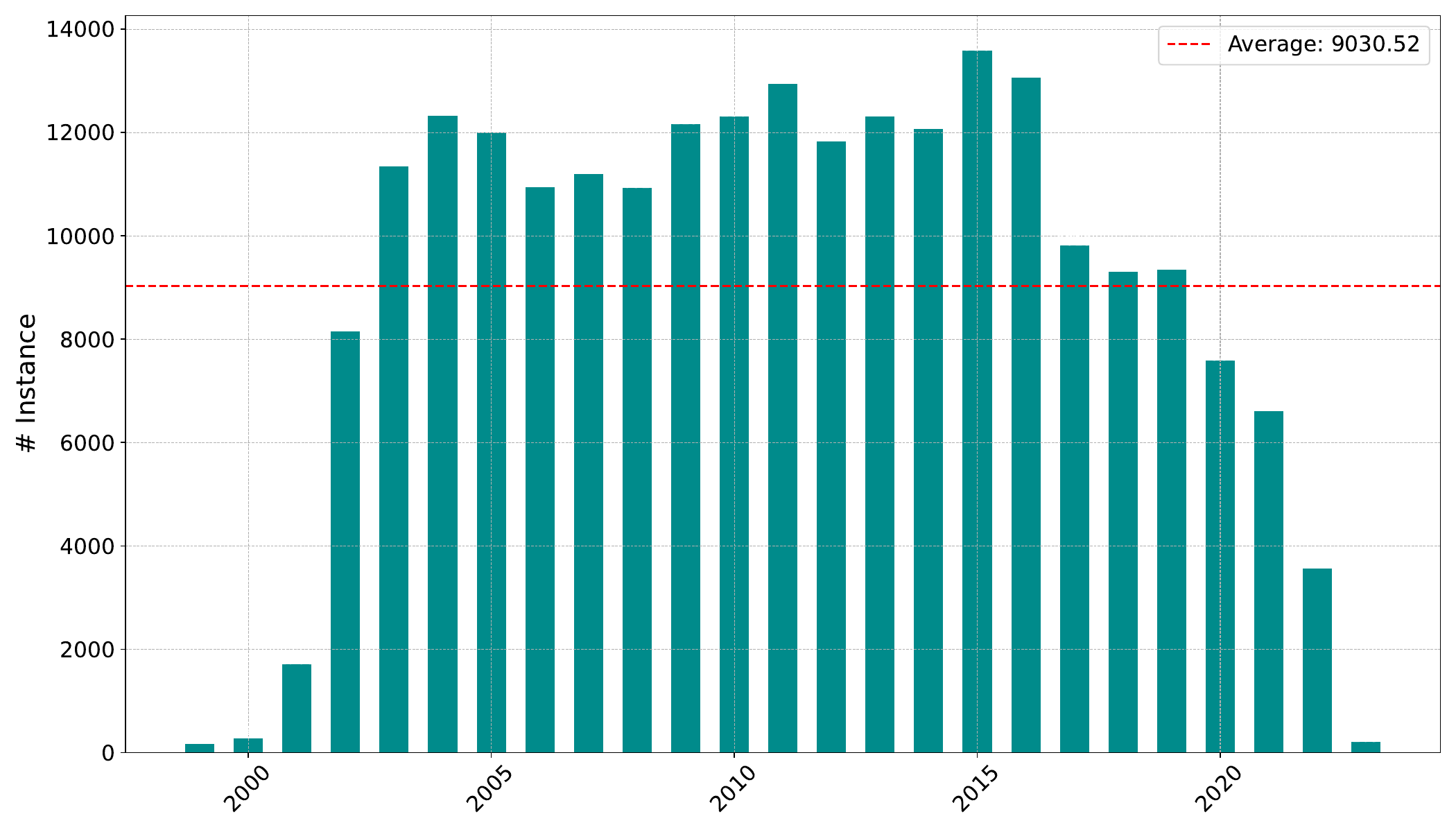}
    \caption{\footnotesize Overview of the years each question was sourced from.}
    \label{fig:count_stats}
\end{figure}

Figure~\ref{fig:count_stats} provides an overview of the years each question was sourced from. The questions spans 25 years starting from 1999 to 2023. The average number of instances for each year is 9030.52, with the most at 2015 by 13583 samples and the least at 1999 with 170 samples. 

\subsection{Copyrights and License}

To prevent copyright issues, we conduct a manual review of the \textit{Test} and \textit{Validation} sets of the KMMLU benchmark. We remove instances from exams with restrictive licenses, primarily in the medical and financial domains. We also filter questions that include segments of Korean literature where the respective authors hold copyright and are challenging to manage individually. However, the \textit{Train} set consists of 208,522 instances, making manual review impractical. Instead, we use insights from our manual review of the test and validation subsets to identify and remove questions in the train set that originate from sources identified as private. Additionally, we collect headers and footers from each webpage during the crawling process to remove those containing copyright information. The final dataset is published under a CC-BY-ND license and is freely available at HuggingFace\footnote{\url{https://huggingface.co/datasets/HAERAE-HUB/KMMLU}}. Additionally during our manual review we do not identify any personally identifiable information or offensive content.

For better reproducibility, our evaluation codebase\footnote{\url{https://github.com/EleutherAI/lm-evaluation-harness/tree/main/lm_eval/tasks/kmmlu}} is built using Eleuther AI's LM-Eval-Harness (MIT License)~\citep{eval-harness}. This includes the evaluation code, generation configuration for all settings (open and proprietary models, direct and CoT settings), and our prompts and CoT exemplars. All components are available under the MIT license.

\newpage
\section{Additional Analysis}
\subsection{How Does Continual Pretraining Affect Korean Proficiency?}~\label{app:contin}
\begin{figure}[ht]
    \centering
    \includegraphics[width=0.6\linewidth]{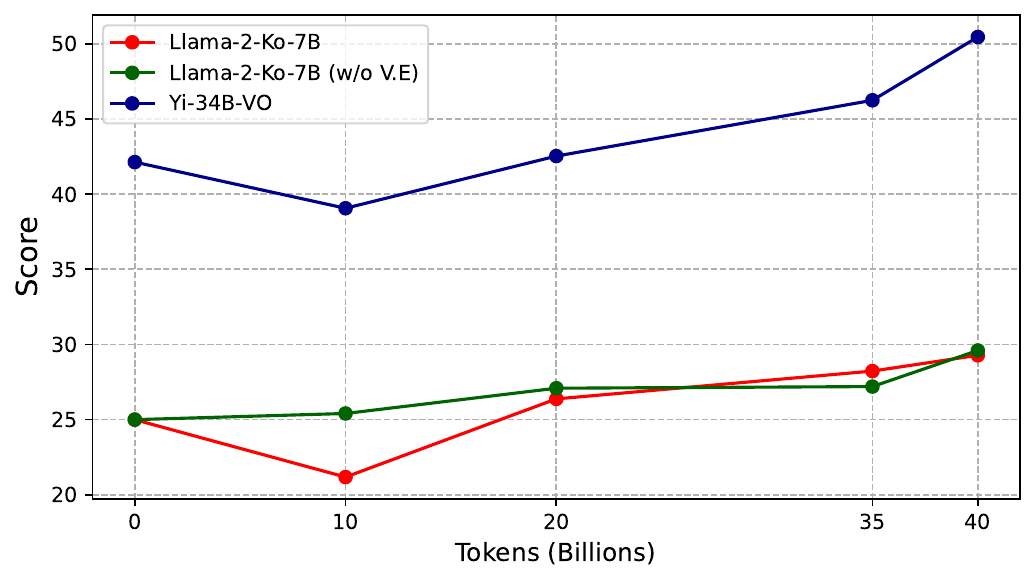}
    \caption{\footnotesize Performance of \textsc{Llama-2-Ko-7B} (both with and without vocabulary expansion) and \textsc{Yi-Ko-34B} on the KMMLU benchmark for each checkpoint.}
    \label{fig:continual_ab}
\end{figure}

In Figure~\ref{fig:continual_ab}, we evaluate the performance of three models—\textsc{Llama-2-Ko-7B} (both with and without vocabulary expansion) and \textsc{Yi-Ko-34B}—across 40 billion tokens of continual pretraining.\footnote{The \textsc{Yi-Ko-6B} model is excluded as intermediary checkpoints are unavailable.} Our analysis reveals two key insights: First, while vocabulary expansion initially leads to a drop in performance, the models subsequently recover and stabilize after surpassing 10 billion tokens of training. This challenges previous findings that suggested vocabulary expansion is unfavorable at such training scales. Interestingly, Table~\ref{tab:perplexity} shows that the perplexity of \textsc{Yi-Ko-34B} is higher than its original model \textsc{Yi-34B}, likely due to the undertraining of newly added tokens, indicating room for improvement. Second, learning outcomes vary despite being trained on the same dataset, suggesting that differences in inherent model capabilities significantly influence performance.


\subsection{Do Machines Handle Problems with Negation Effectively?}

Table~\ref{tab:neg} demonstrates a notable trend in language model performance on the KMMLU test set: models perform better on questions that include negations. This finding contrasts with previous studies~\citep{hosseini-etal-2021-understanding, li2023cmmlu} that identified LLMs to suffer when dealing with negated questions. However, this does not suggest that negation in Korean presents a lower difficulty level than in other languages. Instead, the improved performance may be attributed to the nature of the questions in KMMLU, where negation is more common in declarative knowledge questions, which are generally easier for models to handle compared to procedural knowledge questions~\citep{hendrycks2020measuring}. For example, the math subset, which is the most challenging subset for most LLMs, does not include any negated questions. Furthermore, Table~\ref{tab:negr} illustrates that only 20\% of STEM and 19\% of Applied Science questions include negation, in contrast to 45\% in the HUMSS subset. 

\begin{table}[ht!]
\centering
\fontsize{9}{11}\selectfont
\caption{Comparison of accuracy between questions with and without negation. Evaluation is done in 5-shot setting using the Direct Method.}
\begin{tabular}{lcc}
\toprule
\multicolumn{1}{c}{\textbf{Models}} & \textbf{W Negation} & \textbf{W/O Negation} \\
\midrule
\textsc{Llama-2-70B} & 40.2 & 40.08 \\
\textsc{Yi-34B} & 47.26 & 42.43 \\
\textsc{Qwen-72B} & 53.57 & 48.82 \\
\midrule
\textsc{Gemini-Pro} & 55.05 & 48.63 \\
\textsc{GPT-3.5-Turbo} & 45.61 & 40.39 \\
\textsc{GPT-4} & 65.53 & 57.88 \\ 
\bottomrule
\end{tabular}%
\label{tab:neg}
\end{table}

\begin{table}[ht]
\centering
\fontsize{9}{11}\selectfont
\caption{Ratio of Negated Questions in each category.}
\begin{tabular}{lc}
\toprule
\multicolumn{1}{c}{\textbf{Category}} & \textbf{\% of Negated Q.} \\
\midrule
STEM & 20.54\% \\
Applied Science & 19.16\% \\
HUMSS & 45.76\% \\
Other & 34.83\% \\
\midrule
Math & 0.00\% \\
Electrical Eng. & 9.70\% \\
Aviation Eng. \& Maint. & 14.40\% \\ 
\bottomrule
\end{tabular}%
\label{tab:negr}
\end{table}

\subsection{When Do Korean Proprietary Models Outperform GPT-4?}


\begin{wrapfigure}{r}{0.5\textwidth}
    \vspace{-6mm}
    \centering
    \includegraphics[width=\linewidth]{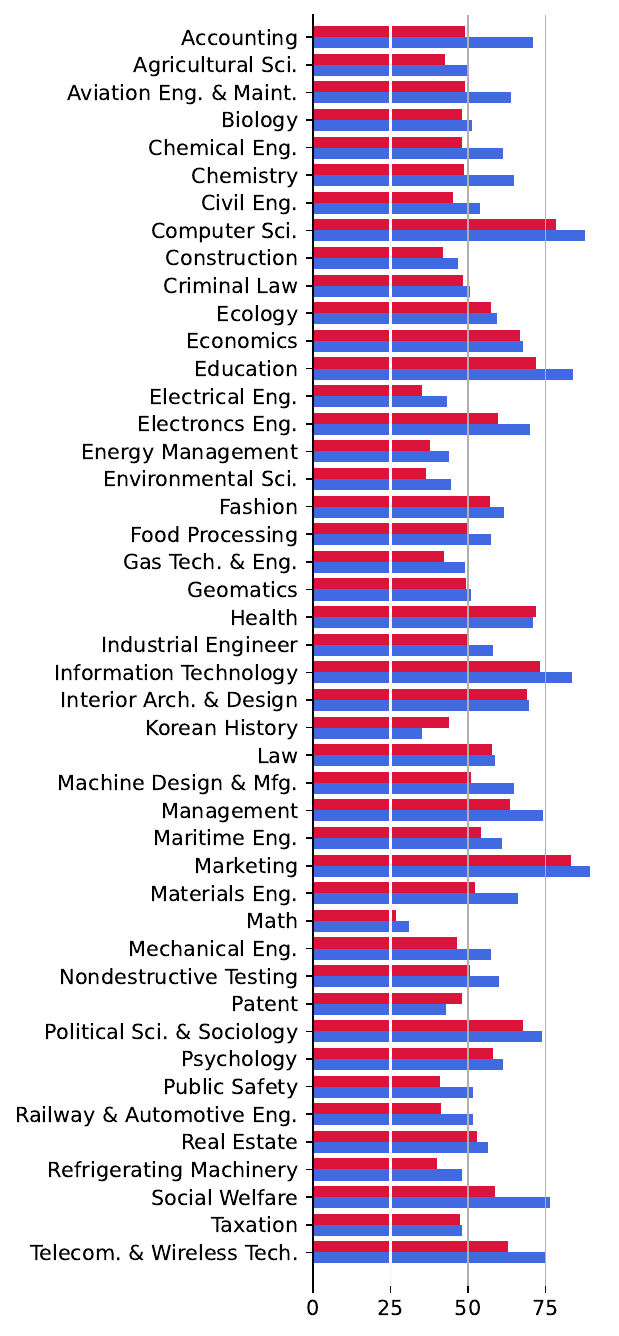}
    \caption{Comparison of \textsc{GPT-4}(Blue) and \textsc{HyperCLOVA X}(Red) using the Direct method in a 5-shot setting.}
    \label{fig:comp}
\vspace{-11mm}
\end{wrapfigure}
Figure~\ref{fig:comp} provides a comparative performance analysis between the top-performing Korean model, \textsc{HyperCLOVA X}, and \textsc{GPT-4} across various disciplines, with detailed numerical results available in Appendix~\ref{fig:comp}. The comparison shows that \textsc{GPT-4} generally outperforms \textsc{HyperCLOVA X} in most subjects, with performance differentials ranging from a significant 22.0\% in Accounting to a marginal 0.5\% in Taxation. Although specific details on the pretraining of both models are kept private \textsc{HyperCLOVA X} is designed for bilingual (English and Korean) use, while \textsc{GPT-4} supports many more languages. This observation corroborates our earlier findings from Section~\ref{sec:curse} that the curse of multilinguality diminish as models scale. Notably, \textsc{HyperCLOVA X} demonstrates superior performance over GPT-4 in Korean History and Criminal Law. This is likely attributable to \textsc{HyperCLOVA X}'s specialized focus on the Korean language, which presumably enhances its proficiency in topics requiring regional-specific knowledge and understanding.

These characteristics are evident in Table~\ref{tab:ko-subset-perf}, which compares the performance of models on the KMMLU-KOR subset. \textsc{HyperCLOVA X}  and \textsc{GPT-4} scores 50.00 and 51.21, respectively. The performance gap narrows as \textsc{GPT-4} experiences a decrease of 8.74 points, while \textsc{HyperCLOVA X} sees a smaller decline of 3.4 points. This indicates that \textsc{HyperCLOVA X} is more resilient to questions on Korean knowledge, maintaining closer to its original performance.

\subsection{Do Machines Also Err Where Humans Often Do?}


\begin{figure}[ht]
    \centering
    \includegraphics[width=0.7\linewidth]{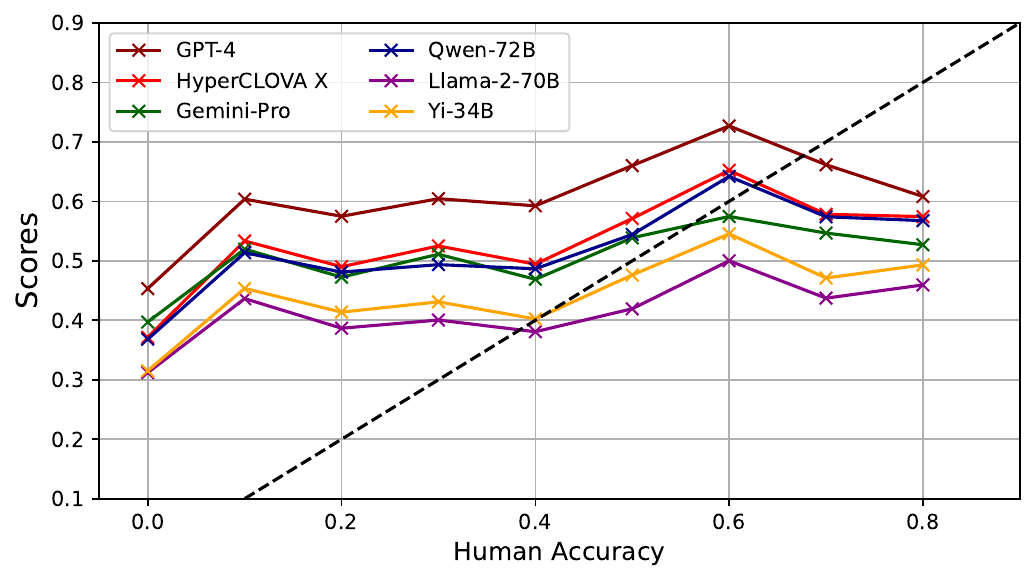}
    \caption{Comparison of model performance and human accuracy. Model performance is calculated using the Direct method in a 5-shot setting.}
    \label{fig:human}
\end{figure}

In Figure~\ref{fig:human}, we compare the performance of LLMs against human accuracy. The findings indicate that LLMs do not exhibit a performance trend that correlates with human performance. Instead, the models display similar performance levels irrespective of the variance in human accuracy. This observation aligns with insights from the \citep{hendrycks2020measuring}, which reported that GPT-3 achieved a higher score in College Mathematics at 35.0\%, compared to 29.9\% in Elementary Mathematics, suggesting that the model's performance does not necessarily scale with the complexity of the task as judged by human standards. Interestingly, the models demonstrate a strong correlation with each other, implying that despite being trained on distinct datasets, they possess similar capabilities. This phenomenon indicates that there may be underlying commonalities in how these models process and generate responses, leading to a similar performance trend.


\section{Contamination Check}\label{app:cont}
The questions for the KMMLU benchmark are sourced from publicly available Korean tests. Accordingly, there may be concerns about whether the sources have been included in the pretraining corpora of the tested language models. Unfortunately, none of the examined models disclose their training data, which precludes a direct assessment of potential knowledge leakage. As an alternative, we refer to the methods of \citet{xu2024benchmarking} to estimate potential leakage using perplexity and n-gram accuracy. We sample 2,000 instances from the KMMLU test set and paraphrase them using \textsc{GPT-4-Turbo}. The paraphrasing is conducted twice to reduce potential bias, with temperatures of 0.7 and 0.8. See Figure~\ref{fig:contam-prompt} for the prompts used.

In Table~\ref{tab:perplexity}, we observe that all models demonstrate lower perplexity on the paraphrased versions of the benchmark. While this does not entirely rule out the possibility of whether the models have encountered the benchmark during pretraining, the preference for a rewritten version—which does not exist online and thus could not have been included in the pretraining—implies that the models are not familiar with the original material.

\begin{table}[ht]
\centering
\fontsize{9}{11}\selectfont
\caption{Comparision of perplexity calculated from the KMMLU Benchmark. The Paraphrased column denotes the average perplexity calculated from the two versions each generated with temperatures 0.7 and 0.8.}
\begin{tabular}{lcc}
\toprule
\textbf{Model} & \textbf{Original} & \textbf{Paraphrased} \\
\midrule
\textsc{Polyglot-Ko-12.8B} & 9.961 & 8.996 \\
\textsc{Yi-34B} & 3.006 & 2.533 \\
\textsc{Yi-Ko-34B} & 12.810 & 9.538 \\
\textsc{Llama-2-70B} & 3.067 & 2.589 \\
\textsc{Qwen-72B} & 10.750 & 7.234 \\
\midrule
\textsc{Yi-34B-Chat} & 3.354 & 2.783 \\
\textsc{Llama-2-70B-Chat} & 5.431 & 3.846 \\
\textsc{Qwen-72B-Chat} & 15.625 & 9.875 \\ 
\bottomrule
\end{tabular}
\label{tab:perplexity}
\end{table}

\begin{figure}[ht]
    \centering
    \includegraphics[width=0.7\linewidth]{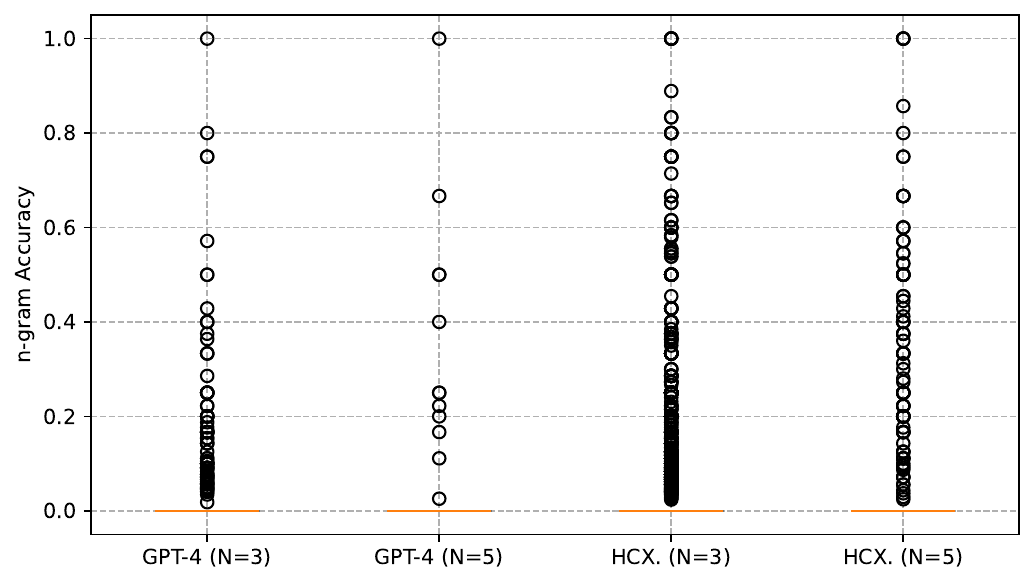}
    \caption{Comparison of model performance and human accuracy. Model performance is calculated using the Direct method in a 5-shot setting.}
    \label{fig:gram}
\end{figure}

For proprietary models where perplexity examinations are unavailable, we use n-gram accuracy. First, we combine the question and answer parts with a single space for each sample, creating a combined text \( X \). Second, we uniformly sample \( K \) (i.e., 5) starting points within the interval from 2 to \(|X|\). The text from the beginning to each starting point serves as the prompt, with the subsequent n-gram used as the prediction target. The prediction results are shown in Figure~\ref{fig:gram}.


\textsc{GPT-4} and \textsc{HyperCLOVA X} demonstrate low average scores for predicting target n-grams, with 0.004 and 0.001 for N=3 and N=5 in \textsc{GPT-4}, and 0.014 and 0.005 for N=3 and N=5 in \textsc{HyperCLOVA X}, respectively. However, \textsc{GPT-4} scores above 0.4 in 21 instances for N=3 and 9 instances for N=5. Examining the samples in Tables~\ref{tab:ngram-qual-0} and~\ref{tab:ngram-qual-1} reveals that the target is a repetition of the prompt, suggesting the model's guesswork rather than data contamination. \textsc{HyperCLOVA X} exhibits a marginally higher average score than \textsc{GPT-4}, scoring above 0.4 in 69 instances for N=3 and 37 instances for N=5. Tables~\ref{tab:ngram-qual-2}, ~\ref{tab:ngram-qual-3}, and ~\ref{tab:ngram-qual-4} present similar findings, with the model occasionally repeating the target verbatim despite its absence from the prompt, potentially indicating leakage. Nonetheless, given the low average scores, we hypothesize that the model might have encountered internet sources reproducing exam segments, rather than contamination of the entire source.

\begin{table}[ht!]
\centering
\fontsize{8}{9}\selectfont
\caption{An example where \textsc{GPT-4} scores high n-gram accuracy. \fbox{fbox} highlights the overlapping n-grams.}
\begin{tabular}{ll}
\toprule
Model & \textsc{GPT-4} \\
\midrule
Prompt & \parbox{0.7\textwidth}{큰날개와 꼬리날개에 의한 무게중심 주위의 키놀이모멘트 관계식은? (단, MC.G : 무게중심 주위의 키놀이 모멘트, MC.G WING : 큰날개에 의한 키놀이 모멘트, MC.G TAIL : 꼬리날개에 의한 키놀이\\}\\
Target & 모멘트)\fbox{MC.G = MC.G WING + MC.G TAIL}\\ \\
Generation & \parbox{0.7\textwidth}{큰날개와 꼬리날개에 의한 무게중심 주위의 키놀이모멘트 관계식은 \fbox{MC.G = MC.G WING + MC.G TAIL} 입니다.\\} \\
Accuracy (N=3, N=5) & (1.0, 0.8) \\
\bottomrule
\end{tabular}

\label{tab:ngram-qual-0}
\end{table}

\begin{table}[ht!]
\centering
\fontsize{8}{9}\selectfont
\caption{An example where \textsc{GPT-4} scores high n-gram accuracy. \fbox{fbox} highlights the overlapping n-grams.}
\begin{tabular}{ll}
\toprule
Model & \textsc{GPT-4} \\
\midrule
Prompt & \parbox{0.7\textwidth}{(주)한공은 종업원 홍길동 씨가 퇴사하여 퇴직금 6,000,000원을 보통예금계좌에서 지급하였다. 홍길동 씨의 퇴사직전 (주)한공의 퇴직급여충당부채는 10,000,000원이다. 홍길동 씨 퇴직금지급에 대한 분개로 옳은 것은?(차) 퇴직급여충당부채\\}\\
Target & : \fbox{6,000,000원, (대) 보통예금 : 6,000,000원}\\ \\
Generation & \parbox{0.7\textwidth}{(주)한공은 종업원 홍길동 씨가 퇴사하여 퇴직금 6,000,000원을 보통예금계좌에서 지급하였다. 홍길동 씨의 퇴사직전 (주)한공의 퇴직급여충당부채는 10,000,000원이다. 홍길동 씨 퇴직금지급에 대한 분개로 옳은 것은?(차) 퇴직급여충당부채(차) 퇴직급여충당부채 \fbox{6,000,000원 (대) 보통예금 6,000,000원}\\} \\
Accuracy (N=3, N=5) & (1.0, 0.8) \\
\bottomrule
\end{tabular}
\label{tab:ngram-qual-1}
\end{table}

\begin{table}[ht!]
\centering
\fontsize{8}{9}\selectfont
\caption{An example where \textsc{HyperCLOVA X} scores high n-gram accuracy}
\begin{tabular}{ll}
\toprule
Model & \textsc{HyperCLOVA X} \\
\midrule
Prompt & \parbox{0.7\textwidth}{큰날개와 꼬리날개에 의한 무게중심 주위의 키놀이모멘트 관계식은? (단, MC.G : 무게중심 주위의 키놀이 모멘트, MC.G WING : 큰날개에 의한 키놀이 모멘트, MC.G TAIL : 꼬리날개에 의한 키놀이\\}\\
Target & 모멘트)\fbox{MC.G = MC.G WING + MC.G TAIL}\\ \\
Generation & \parbox{0.7\textwidth}{무게중심 주위의 키놀이 모멘트(MC.G)는 큰날개에 의한 키놀이 모멘트(MC.G WING)와 꼬리날개에 의한 키놀이 모멘트(MC.G TAIL)의 합으로 구할 수 있습니다.

- \fbox{MC.G = MC.G WING + MC.G TAIL}\\} \\
Accuracy (N=3, N=5) & (1.0, 0.8) \\
\bottomrule
\end{tabular}
\label{tab:ngram-qual-2}
\end{table}

\begin{table}[ht!]
\centering
\fontsize{8}{9}\selectfont
\caption{An example where \textsc{HyperCLOVA X} scores high n-gram accuracy}
\begin{tabular}{ll}
\toprule
Model & \textsc{HyperCLOVA X} \\
\midrule
Prompt & \parbox{0.7\textwidth}{기계식주차장의 기준에 관한 기술이 잘못 된 것은?대형기계식주차장은 직경 4m 이상의 방향전환장치와 그 방향전환장치에\\}\\
Target &  \fbox{접한 너비 1m 이상의 여유공지가 있어야 한다.}\\ \\
Generation & \parbox{0.7\textwidth}{\fbox{접한 너비 1m 이상의 여유공지가 있어야 합니다.}\\}\\
Accuracy (N=3, N=5) & (1.0, 1.0) \\
\bottomrule
\end{tabular}
\label{tab:ngram-qual-3}
\end{table}

\begin{table}[ht!]
\centering
\fontsize{8}{9}\selectfont
\caption{An example where \textsc{HyperCLOVA X} scores high n-gram accuracy}
\begin{tabular}{ll}
\toprule
Model & \textsc{HyperCLOVA X} \\
\midrule
Prompt & \parbox{0.7\textwidth}{철골콘크리트조의 기둥에서 내화구조의 기준으로 옳은 것은?작은지름 25cm 이상으로서 철골을\\}\\
Target & \fbox{두께 5cm 이상의 콘크리트로 덮은 것}\\ \\
Generation & \parbox{0.7\textwidth}{철골콘크리트조의 기둥에서 내화구조의 기준으로 옳은 것은?작은지름 25cm 이상으로서 철골을철골철근콘크리트조의 기둥에서 내화구조의 기준은 다음과 같습니다... \\
4. 철골보를 \fbox{두께 5cm 이상의 콘크리트로 덮은 것}...\\}\\
Accuracy (N=3, N=5) & (1.0, 1.0) \\
\bottomrule
\end{tabular}
\label{tab:ngram-qual-4}
\end{table}

\clearpage
\section{License}~\label{app:license}
The KMMLU benchmark is released under a CC-BY-ND license. This license prohibits the distribution of the remixed or transformed version of the dataset. The code and prompts for evaluation is released by a MIT license via LM-Eval-Harness.


\section{Evaluated Models}~\label{app:models}
\textbf{Polyglot-Ko}~\cite{ko2023polyglot}. Introduced by the Polyglot Team of EleutherAI \textsc{Polyglot-Ko} is a comprehensive suite of Korean-centric autoregressive language models featuring models with 1.3, 3.8, 5.8, and 12.8 billion parameters. The models are pretrained on Korean corpus ranging from 167 to 219 billion tokens.

\noindent \textbf{Llama-2}~\citep{touvron2023llama}. \textsc{Llama-2} is a suite of large language models ranging from 7 to 70 billion parameters developed by Meta. The models are pretrained on 2 trillion tokens, and whether Korean is included is not reported. The suite also provides \textsc{Llama-2-Chat} an aligned version for instruction-following and ssafety.

\noindent\textbf{Yi}~\citep{yi}. The \textsc{Yi} model, developed by 01.AI, is a series of bilingual language models available in two variants: 6B and 34B. It employs an architecture similar to \textsc{Llama-2} and is pretrained on a multilingual corpus of 3 trillion tokens. Additionally, the model features chat versions tailored for instruction-following.

\noindent\textbf{Qwen}~\citep{bai2023qwen}. \textsc{Qwen} is a suite of bilingual language models developed by Alibaba Cloud, with variants spanning from 1.8 billion to 72 billion parameters. Each model within the series is pretrained on a dataset of 3 trillion tokens. The \textsc{Qwen} also includes specialized chat models designed for following instructions.

\noindent \textbf{GPT-3.5 \& GPT-4}~\citep{openai2023gpt4}. Developed by OpenAI, the \textsc{GPT} series is renowned for exhibiting state-of-the-art performance across various benchmarks and tasks, including exceptional instruction-following capabilities. Specific details regarding the parameter count and the scope of the training data are not open to the public.

\noindent \textbf{Gemini}~\citep{team2023gemini}. \textsc{Gemini} is a series of models developed by Google, encompassing four variants: Nano-1, Nano-2, Pro, and Ultra. In our experiments, we utilize \textsc{Gemini-Pro}. Details regarding the parameter count and the dataset used for training are not disclosed.

\noindent\textbf{HyperCLOVA X}~\citep{kim2021changes}. \textsc{HyperCLOVA X}, developed by NAVER, is a bilingual language model proficient in both English and Korean.

\begin{table}[ht!]
\centering
\fontsize{9}{11}\selectfont
\caption{Overview of the 31 LLMs evaluated in this paper.}
\begin{tabular}{lccc}
\toprule
\textbf{Model} & \textbf{\# Params} & \textbf{Access} & \textbf{Language} \\
\midrule
\multicolumn{4}{c}{\textit{\textbf{English-Centric / Bilingual Pretrained Models}}} \\
\midrule
\textsc{Llama-2}~\citep{touvron2023llama} & 7B, 13B, 70B & Weights Available & En \\
\textsc{Yi}~\citep{yi} & 6B, 34B & Weights Available & En / Zh \\
\textsc{Qwen}~\citep{bai2023qwen} & 7B, 14B, 72B & Weights Available & En / Zh \\
\midrule
\multicolumn{4}{c}{\textit{\textbf{English-Centric / Bilingual Finetuned Models}}} \\
\midrule
\textsc{Llama-2-Chat}~\citep{touvron2023llama} & 7B, 13B, 70B & Weights Available & En \\
\textsc{Yi-Chat}~\citep{yi} & 6B, 34B & Weights Available & En / Zh \\
\textsc{Qwen-Chat}~\citep{bai2023qwen} & 7B, 14B,72B & Weights Available & En / Zh \\
\midrule
\multicolumn{4}{c}{\textit{\textbf{Korean Pretrained Models}}} \\
\midrule
\textsc{Polyglot-Ko}~\citep{ko2023polyglot} & 1.3B, 3.8B, 5.8B, 12.8B & Open Source & Ko \\
\midrule
\multicolumn{4}{c}{\textit{\textbf{Korean Continual Pretrained Models}}} \\
\midrule
\textsc{Llama-2-Ko}~\citep{l._junbum_2023} & 7B & Weights Available & En / Ko \\
\textsc{Yi-Ko}~\citep{l._junbum_2023} & 6B, 34B & Weights Available & En / Zh / Ko \\
\midrule
\multicolumn{4}{c}{\textit{\textbf{Proprietary Models}}} \\
\midrule
\textsc{GPT-3.5-Turbo} & undisclosed & API & - \\
\textsc{GPT-4}~\citep{openai2023gpt4} & undisclosed & API & - \\
\textsc{Gemini-Pro}~\citep{team2023gemini}  & undisclosed & API & - \\
\textsc{HyperCLOVA X}~\cite{kim2021changes} & undisclosed & API & - \\ 
\bottomrule
\end{tabular}%
\label{tab:models}
\end{table}

\section{Compute Resources}\label{app:compute}
Models with openly available weights were evaluated on an internal cluster comprising 4 NVIDIA A100 HBM2e 80GB PCIe GPUs. We performed 44 direct evaluations on the original KMMLU set and 2 CoT evaluations on the KMMLU-HARD set, totaling approximately 300 A100 GPU hours.

\section{Prompting Format}
For evaluation, we use the following prompting formats. For n-shot evaluation, the identical format is repeated.

\begin{figure}[ht!]
\fontsize{8}{9}\selectfont
    \centering
    \begin{tabular}{|p{0.9\linewidth}|}
    \hline 
    \\
    \textbf{Direct Evaluation Prompt} \\
         \\
         \{Question\} \\
         A. \{A\} \\
         B. \{B\} \\
         C. \{C\} \\
         D. \{D\} \\ 
         정답： \\
\\\\
\hline
    \end{tabular}
    \caption{Prompt used in our Direct Evaluation.}
    \label{fig:direct-prompt}
\end{figure}

\begin{figure}[ht!]
\fontsize{8}{9}\selectfont
    \centering
    \begin{tabular}{|p{0.9\linewidth}|}
    \hline 
    \\
    \textbf{CoT Evaluation Prompt} \\
         \\
        질문: \{Question\} \\ 
        A. \{A\} \\
        B. \{B\} \\
        C. \{C\} \\
        D. \{D\} \\ 
        정답: 차근 차근 생각해봅시다. 회계학 관련 정보를 위해 위키피디아를 참조하겠습니다.\\
\\\\
\hline
    \end{tabular}
    \caption{Prompt used in our CoT Evaluation.}
    \label{fig:cot-prompt}
\end{figure}

For the paraphrasing in Section~\ref{app:cont}, we use the following prompting format.

\begin{figure}[ht!]
\fontsize{8}{9}\selectfont
    \centering
    \begin{tabular}{|p{0.9\linewidth}|}
    \hline 
    \\
You are a problem rewriter. Your task is to paraphrase the problem and the answer presented below.\\
\\
Please follow the instructions below:\\
\\
1. Please paraphrase the problem by rewording it with new expressions and sentence structures.\\
2. Please do not change the essence of the problem and the answer.\\
3. Please make sure not to deviate too much from the original content, and maintain the style as much as possible. \\
4. Please write in coherent Korean. It should sound natural to a native Korean speaker.\\
\\
Please write ”Rewritten Question: <question>” to output your rewritten question without any additional information, and write ”Rewritten Answer: <answer>” to output your rewritten answer without any additional information.\\
\\
There is an example for your reference:\\
\\
Original Question: (주)한국은 20× 1년 1월 초 A사 지분 상품을 ￦ 10,000 에 매입하면서 매입 수수료 ￦ 500을 현금으로 지급하고, 기타 포괄손익 － 공정가치 측정 금융자산으로 분류하였다. 20× 1년 12월 말 A사 지분 상품의 공정가치가 ￦8,000이라면, 20× 1년 말 (주)한국이 인식할 A사 지분 상품 관련 평가손익은?\\
Original Answer: 금융자산 평가손실 (기타 포괄손익 ) ￦2,500\\
\\
Rewritten Question: 한국 회사가 20× 1년 1월에 A사 지분 상품을 ￦ 10,000에 구매하여 현금으로 ￦ 500의 매입 수수료를 내고, 이를 기타 포괄손익 - 공정가치 측정 금융자산으로 분류했습니다. 20× 1년 12월 말에 A사 지분 상품의 공정가치가 ￦8,000이라면, 20× 1년 말에 한국 회사가 A사 지분 상품과 관련된 평가손익은 얼마입니까?\\
Rewritten Answer: 한국 회사의 금융자산 평가손실은 ￦2,500입니다.\\
\\
Original Question: \{Question\}\\
Original Answer: \{Answer\}\\

\\\\
\hline
    \end{tabular}
    \caption{Prompt used in paraphrasing for contamination check.}
    \label{fig:contam-prompt}
\end{figure}

\section{More details for CoT Exemplar Creation}
\label{appendix:cot-exemplar-creation}

\begin{figure}[ht!]
\fontsize{8}{9}\selectfont
    \centering
    \begin{tabular}{|p{0.9\linewidth}|}
    \hline 
    \\
    \textbf{CoT Elicitation Prompt} \\
         \\
         다음은 \{category\}에 대한 객관식 질문입니다. \\
         정확한 답을 하기 위해 반드시 웹 브라우징을 활용하시오. 먼저 자세한 정답 추론/해설 과정을 한글로 생성하세요. \\
         그리고나서, 최종 답변은 반드시 다음과 같은 포맷으로 답해야 합니다. '따라서, 정답은 (A|B|C|D)입니다.' \\
         \\\\
         질문: \{Question\} \\
         선택지: \\
         (A). \{A\} \\
         (B). \{B\} \\
         (C). \{C\} \\
         (D). \{D\} \\
         정답 해설: 차근 차근 생각해보겠습니다. \\
\\\\
\hline
    \end{tabular}
    \caption{Zero-shot CoT prompt used in our CoT exemplar creation.}
    \label{fig:cot-elicit-prompt}
\end{figure}

We use the zero-shot CoT prompt of Figure~\ref{fig:cot-elicit-prompt} to collect the exemplar CoTs for our dataset. We request to use browsing for more accurate explanations if it is available. For GPT-4, we manually input the prompt to the ChatGPT Web interface (\href{https://chat.openai.com}{chat.openai.com}). For HyperCLOVA X, we devise 3-shot demonstrations to generate relevant queries to the NAVER search engine (\href{https://www.naver.com/}{www.naver.com}). Then, we concatenate top-3 search results to generate explanations.


\section{KMMLU-KOR Subset}~\label{app:kor}
For more targeted evaluation, alongside KMMLU-HARD, we introduce a second subset, KMMLU-KR. KMMLU-KR is a collection of questions that specifically require Korean knowledge to solve so that users can assess a language model's proficiency in Korea-specific contexts. We initially adopted a keyword filter to collect 2,149 candidate questions. Then, two authors independently reviewed each question, eliminating irrelevant entries and categorizing the pertinent ones into four distinct categories: \textit{Cultural}, which delves into Korean history and societal norms;  \textit{Regional}, focusing on the geographical details of Korea; \textit{Legal}, concerning Korea's legal and governmental structures; \textit{Others}, comprising questions that demand knowledge of the Korean context but do not fit into the previous categories. Following the filtering, 1,305 questions remained, constituting the KMMLU-KOR subset. Please refer to Figure~\ref{fig:examples} for detailed examples of each category. For evaluation results, see Table~\ref{tab:ko-subset-perf}.


\begin{table}[ht!]
\fontsize{9}{11}\selectfont
\centering
\caption{\footnotesize Average accuracy(\%) calculated using the Direct method in a 5-shot setting on the KMMLU-KOR subset. We report the micro-average accuracy due to the imbalance on each category. The highest-scoring model across the entire table is highlighted in \textbf{bold}, and the best model within each category is \underline{underlined}.}
\begin{tabular}{lccccc}
\toprule
\multicolumn{1}{c}{\textbf{Model}} & \textbf{Cultural} & \textbf{Legal} & \textbf{Regional} & \textbf{Other} & \textbf{Total} \\
\midrule
\multicolumn{6}{c}{\textit{\textbf{Multilingual Pretrained Models}}} \\
\midrule
\textsc{Llama-2-7B} & 21.51 & 22.77 & 38.46 & 25.53 & 23.26 \\
\textsc{Llama-2-13B} & 29.34 & 27.16 & 50.00 & 28.87 & 28.13 \\
\textsc{Llama-2-70B} & 35.33 & 34.05 & 30.77 & 40.14 & 34.88 \\
\textsc{Yi-6B} & 31.33 & 33.01 & \underline{50.00} & 33.58 & 33.13 \\
\textsc{Yi-34B} & 37.35 & 41.02 & 38.46 & \underline{47.01} & 41.18 \\
\textsc{Qwen-7B} & 10.47 & 10.08 & 26.92 & 21.28 & 11.63 \\
\textsc{Qwen-14B} & 22.16 & 21.00 & 34.62 & 26.76 & 22.04 \\
\textsc{Qwen-72B} & \underline{37.95} & \underline{43.20} & 34.62 & 46.27 & \underline{42.56} \\
\midrule
\multicolumn{6}{c}{\textit{\textbf{Multilingual Chat Models}}} \\
\midrule
\textsc{Llama-2-7B-Chat} & 23.84 & 26.72 & 42.31 & 28.37 & 26.78 \\
\textsc{Llama-2-13B-Chat} & 27.11 & 24.76 & 46.15 & 29.10 & 25.95 \\
\textsc{Llama-2-70B-Chat} & 27.71 & 30.83 & 53.85 & 29.85 & 30.62 \\
\textsc{Yi-6B-Chat} & 25.58 & 32.85 & \underline{42.31} & 32.62 & 31.98 \\
\textsc{Yi-34B-Chat} & \underline{37.95} & \underline{41.02} & \underline{42.31} & \underline{45.52} & \underline{41.09} \\
\textsc{Qwen-7B-Chat} & 6.36 & 8.97 & 26.92 & 16.78 & 9.81 \\
\textsc{Qwen-14B-Chat} & 20.36 & 17.56 & 34.62 & 26.76 & 19.34 \\
\textsc{Qwen-72B-Chat} & 34.10 & 35.27 & 30.77 & 37.58 & 35.13 \\
\midrule
\multicolumn{6}{c}{\textit{\textbf{Korean Pretrained Models}}} \\
\midrule
\textsc{Polyglot-Ko-1.3B} & \underline{30.81} & \underline{29.00} & \underline{34.62} & \underline{26.24} & \underline{29.00} \\
\textsc{Polyglot-Ko-3.8B} & 25.58 & 25.88 & 26.92 & 21.99 & 25.48 \\
\textsc{Polyglot-Ko-3.8B} & 24.10 & 27.79 & 19.23 & 26.12 & 26.90 \\
\textsc{Polyglot-Ko-12.8B} & 30.12 & 28.03 & 26.92 & 24.63 & 27.85 \\
\midrule
\multicolumn{6}{c}{\textit{\textbf{Korean Continual Pretrained Models}}} \\
\midrule
\textsc{Yi-Ko-6B} & 34.94 & 36.65 & 30.77 & 44.78 & 37.11 \\
\textsc{Yi-Ko-34B} & \underline{44.19} & \underline{44.70} & \underline{42.31} & \underline{51.77} & \underline{45.14} \\
\midrule
\multicolumn{6}{c}{\textit{\textbf{Proprietary Models}}} \\
\midrule
\textsc{GPT-3.5-Turbo} & 38.73 & 34.15 & 26.92 & 41.61 & 35.36 \\
\textsc{Gemini-Pro} & \underline{45.18} & 39.68 & \underline{42.31} & 43.28 & 40.83 \\
\textsc{HyperCLOVA X} & \textbf{47.90} & \underline{50.65} & \underline{42.31} & \underline{51.41} & \underline{50.00} \\
\textsc{GPT-4} & 41.57 & \textbf{52.31} & \textbf{46.15} & \textbf{58.21} & \textbf{51.21} \\
\bottomrule
\end{tabular}
\label{tab:ko-subset-perf}
\end{table}

\section{Evaluation Results}~\label{app:results}
In this section, we present the results of our evaluation, broken down by category for each model assessed. Tables~\ref{tab:ap_ko_p}-\ref{tab:ap_pro} include results using the Direct method. Table~\ref{tab:ap_cot} presents the results evaluated using the CoT method. Figure~\ref{fig:comp} presents a comparative performance
analysis between the most capable Korean model, \textsc{HyperCLOVA X}, and \textsc{GPT-4}.

\begin{table}[ht!]
\fontsize{9}{11}\selectfont
\centering
\caption{5-shot accuracy using the Direct method for  \textsc{Polyglot-Ko}, and  \textsc{Yi-Ko} broken down by category.}
\begin{tabular}{lcccccc}
\toprule
\textbf{Category} & \multicolumn{4}{c}{\textsc{Polyglot-Ko}} & \multicolumn{2}{c}{\textsc{Yi-Ko}}\\
 & 1.3B & 3.8B & 5.8B & 12.8B & 6B & 34B \\
\midrule
accounting & 30.0 & 32.0 & 32.0 & 30.0 & 38.0 & 46.0 \\
agricultural\_sciences & 27.0 & 30.3 & 30.1 & 32.0 & 32.7 & 39.6 \\
aviation\_engineering\_and\_maintenance & 30.2 & 29.7 & 29.9 & 30.7 & 36.1 & 45.2 \\
biology & 24.0 & 26.7 & 28.6 & 25.3 & 32.1 & 41.7 \\
chemical\_engineering & 25.3 & 27.9 & 24.7 & 24.7 & 36.2 & 49.7 \\
chemistry & 30.3 & 25.2 & 26.0 & 29.2 & 40.8 & 52.33 \\
civil\_engineering & 27.4 & 31.8 & 31.9 & 34.3 & 38.0 & 45.3 \\
computer\_science & 32.1 & 35.9 & 34.8 & 33.9 & 61.5 & 73.9 \\
construction & 33.6 & 31.0 & 31.7 & 32.0 & 34.7 & 38.3 \\
criminal\_law & 26.0 & 29.0 & 29.5 & 28.5 & 31.5 & 37.0 \\
ecology & 28.7 & 29.4 & 31.8 & 32.7 & 45.2 & 57.4 \\
economics & 23.8 & 26.2 & 24.6 & 24.6 & 41.5 & 58.5 \\
education & 23.0 & 20.0 & 24.0 & 25.0 & 53.0 & 71.0 \\
electrical\_engineering & 29.3 & 32.5 & 32.0 & 32.6 & 34.9 & 36.4 \\
electronics\_engineering & 30.5 & 30.0 & 35.2 & 33.3 & 47.1 & 56.6 \\
energy\_management & 28.8 & 26.5 & 24.5 & 26.9 & 30.0 & 38.4 \\
environmental\_science & 26.1 & 32.9 & 27.3 & 30.9 & 33.9 & 40.8 \\
fashion & 27.0 & 29.5 & 29.2 & 29.8 & 46.1 & 50.9 \\
food\_processing & 27.3 & 31.8 & 33.5 & 29.4 & 36.1 & 45.8 \\
gas\_technology\_and\_engineering & 31.9 & 30.9 & 30.2 & 30.9 & 32.5 & 38.5 \\
geomatics & 29.2 & 30.0 & 31.1 & 31.0 & 41.6 & 46.9 \\
health & 26.0 & 32.0 & 27.0 & 25.0 & 52.0 & 73.0 \\
industrial\_engineer & 27.4 & 32.3 & 33.1 & 31.2 & 43.0 & 48.7 \\
information\_technology & 34.2 & 34.1 & 34.0 & 30.8 & 57.1 & 70.9 \\
interior\_architecture\_and\_design & 32.4 & 29.6 & 29.7 & 31.8 & 47.3 & 61.1 \\
korean\_history & 34.0 & 26.0 & 25.0 & 31.0 & 33.0 & 42.0 \\
law & 26.0 & 24.2 & 24.4 & 23.9 & 41.8 & 53.6 \\
machine\_design\_and\_manufacturing & 28.7 & 34.0 & 26.9 & 30.3 & 39.9 & 45.9 \\
management & 27.6 & 27.7 & 27.7 & 28.0 & 43.7 & 61.9 \\
maritime\_engineering & 24.8 & 31.5 & 26.7 & 26.5 & 44.0 &51.8 \\
marketing & 24.4 & 30.6 & 26.4 & 33.5 & 69.6 & 81.2 \\
materials\_engineering & 30.9 & 30.2 & 30.1 & 26.9 & 39.8 & 51.8 \\
math & 30.0 & 21.3 & 20.0 & 24.7 & 24.0 & 31.3 \\
mechanical\_engineering & 24.2 & 31.5 & 27.1 & 26.9 & 38.0 & 44.1 \\
nondestructive\_testing & 26.4 & 32.1 & 34.2 & 30.3 & 39.0 & 50.6 \\
patent & 29.0 & 23.0 & 22.0 & 31.0 & 32.0 & 34.0 \\
political\_science\_and\_sociology & 25.7 & 25.7 & 25.7 & 25.7 & 41.7 & 64.7 \\
psychology & 26.5 & 25.9 & 27.7 & 25.9 & 40.1 & 53.8\\
public\_safety & 28.5 & 30.7 & 31.5 & 31.3 & 32.1 & 40.6 \\
railway\_and\_automotive\_engineering & 23.6 & 29.0 & 28.9 & 26.8 & 34.7 & 39.6 \\
real\_estate & 27.0 & 27.5 & 29.5 & 32.0 & 45.0 & 60.0 \\
refrigerating\_machinery & 27.0 & 28.9 & 29.7 & 28.3 & 30.0 & 37.9 \\
social\_welfare & 25.3 & 28.9 & 30.0 & 28.8 & 44.7 & 58.5 \\
taxation & 29.0 & 27.0 & 23.5 & 26.5 & 36.5 & 43.0 \\
telecommunications\_and\_wireless\_technology & 28.6 & 33.9 & 34.1 & 32.2 & 52.4 & 60.4 \\ 
\bottomrule
\end{tabular}%

\label{tab:ap_ko_p}
\end{table}

\begin{table}[ht!]
\fontsize{9}{11}\selectfont
\centering
\caption{5-shot accuracy using the Direct method for  \textsc{Llama-2} (original and chat versions) broken down by category.}
\begin{tabular}{lcccccc}
\toprule
\textbf{Category} & \multicolumn{2}{c}{LLama-2-7B} & \multicolumn{2}{c}{LLama-2-13B} & \multicolumn{2}{c}{LLama-2-70B} \\
 & Org. & Chat & Org. & Chat & Org. & Chat \\
 \midrule
accounting & 25.0 & 22.0 & 20.0 & 16.0 & 34.0 & 26.0 \\
agricultural\_sciences & 23.7 & 31.0 & 29.6 & 27.4 & 33.6 & 32.7 \\
aviation\_engineering\_and\_maintenance & 23.7 & 26.8 & 30.3 & 26.8 & 35.9 & 33.0 \\
biology & 23.6 & 26.4 & 28.8 & 25.2 & 33.0 & 28.1 \\
chemical\_engineering & 27.0 & 28.5 & 32.7 & 31.3 & 38.5 & 33.1 \\
chemistry & 26.8 & 26.7 & 30.3 & 27.7 & 41.8 & 32.3 \\
civil\_engineering & 26.9 & 32.1 & 33.8 & 31.1 & 36.4 & 35.4 \\
computer\_science & 24.1 & 28.0 & 47.4 & 41.5 & 67.3 & 58.9 \\
construction & 22.9 & 31.3 & 30.1 & 28.2 & 31.8 & 33.6 \\
criminal\_law & 26.5 & 26.5 & 30.0 & 22.0 & 30.0 & 25.0 \\
ecology & 16.8 & 28.0 & 32.5 & 31.0 & 43.7 & 38.7 \\
economics & 27.7 & 30.8 & 27.7 & 38.5 & 45.4 & 40.0 \\
education & 24.0 & 29.0 & 26.0 & 28.0 & 56.0 & 38.0 \\
electrical\_engineering & 27.4 & 29.4 & 34.0 & 28.0 & 30.8 & 32.3 \\
electronics\_engineering & 33.0 & 32.2 & 38.8 & 31.5 & 47.1 & 39.9 \\
energy\_management & 23.5 & 25.4 & 26.6 & 24.8 & 30.8 & 28.9 \\
environmental\_science & 27.5 & 30.4 & 32.9 & 29.0 & 28.3 & 29.6 \\
fashion & 27.8 & 30.0 & 32.2 & 32.4 & 41.8 & 36.2 \\
food\_processing & 17.4 & 24.3 & 31.1 & 26.6 & 33.9 & 29.9 \\
gas\_technology\_and\_engineering & 22.3 & 28.0 & 29.1 & 26.4 & 31.4 & 29.6 \\
geomatics & 26.9 & 31.0 & 35.4 & 30.5 & 40.2 & 36.9 \\
health & 22.0 & 21.0 & 30.0 & 25.0 & 53.0 & 42.0 \\
industrial\_engineer & 24.5 & 28.9 & 36.5 & 34.3 & 41.9 & 38.6 \\
information\_technology & 27.3 & 29.3 & 44.4 & 37.3 & 62.8 & 52.0 \\
interior\_architecture\_and\_design & 28.3 & 30.2 & 36.0 & 33.0 & 47.8 & 40.8 \\
korean\_history & 26.0 & 21.0 & 25.0 & 25.0 & 32.0 & 23.0 \\
law & 24.4 & 25.5 & 26.5 & 27.6 & 40.8 & 34.9 \\
machine\_design\_and\_manufacturing & 24.0 & 29.2 & 34.1 & 27.9 & 41.8 & 35.0 \\
management & 24.0 & 25.5 & 29.7 & 27.1 & 47.8 & 37.2 \\
maritime\_engineering & 30.0 & 30.3 & 32.8 & 29.7 & 40.3 & 34.7 \\
marketing & 24.0 & 25.1 & 38.7 & 37.2 & 70.7 & 57.4 \\
materials\_engineering & 21.2 & 28.5 & 29.0 & 26.2 & 40.4 & 30.8 \\
math & 25.0 & 28.3 & 24.3 & 26.7 & 27.0 & 23.7 \\
mechanical\_engineering & 25.3 & 29.4 & 34.6 & 28.0 & 31.0 & 30.5 \\
nondestructive\_testing & 24.8 & 29.9 & 34.2 & 25.8 & 41.5 & 32.1 \\
patent & 25.0 & 24.0 & 26.0 & 26.0 & 33.0 & 25.0 \\
political\_science\_and\_sociology & 23.7 & 27.7 & 25.3 & 30.7 & 47.3 & 36.0 \\
psychology & 24.7 & 24.9 & 25.2 & 23.5 & 39.1 & 28.0 \\
public\_safety & 28.5 & 30.4 & 32.6 & 31.0 & 33.0 & 34.0 \\
railway\_and\_automotive\_engineering & 22.7 & 26.7 & 31.2 & 27.3 & 32.4 & 30.0 \\
real\_estate & 23.5 & 24.5 & 24.5 & 25.0 & 32.0 & 26.5 \\
refrigerating\_machinery & 24.2 & 26.3 & 28.7 & 27.8 & 30.1 & 30.8 \\
social\_welfare & 26.7 & 28.8 & 31.9 & 27.6 & 47.8 & 35.0 \\
taxation & 23.0 & 24.5 & 21.5 & 24.5 & 33.0 & 31.0 \\
telecommunications\_and\_wireless\_technology & 27.9 & 29.5 & 44.4 & 35.1 & 54.2 & 44.0\\
\bottomrule
\end{tabular}%

\label{tab:ap_llama}
\end{table}

\begin{table}[]
\fontsize{9}{11}\selectfont
\centering
\caption{5-shot accuracy using the Direct method for  \textsc{Yi} (original and chat versions) broken down by category.}
\begin{tabular}{lcccc}
\toprule
\textbf{Category} & \multicolumn{2}{c}{\textsc{Yi-6B}} & \multicolumn{2}{c}{\textsc{Yi-34B}} \\
 & Org. & Chat & Org. & Chat \\
\midrule
accounting & 29.0 & 30.0 & 46.0 & 45.0 \\
agricultural\_sciences & 32.7 & 29.5 & 36.0 & 34.7 \\
aviation\_engineering\_and\_maintenance & 31.9 & 30.9 & 36.9 & 34.8 \\
biology & 28.9 & 29.4 & 32.5 & 30.9 \\
chemical\_engineering & 31.8 & 31.5 & 40.8 & 40.7 \\
chemistry & 36.7 & 35.0 & 47.5 & 42.3 \\
civil\_engineering & 32.8 & 33.1 & 40.9 & 36.9 \\
computer\_science & 54.0 & 56.8 & 72.1 & 72.0 \\
construction & 30.9 & 30.9 & 34.7 & 30.4 \\
criminal\_law & 34.5 & 36.5 & 39.0 & 37.5 \\
ecology & 34.3 & 35.1 & 46.7 & 44.5 \\
economics & 36.9 & 36.9 & 43.1 & 48.5 \\
education & 40.0 & 44.0 & 58.0 & 62.0 \\
electrical\_engineering & 33.0 & 31.5 & 33.3 & 28.4 \\
electronics\_engineering & 41.9 & 43.2 & 50.4 & 50.1 \\
energy\_management & 28.8 & 30.5 & 33.8 & 32.7 \\
environmental\_science & 31.5 & 29.5 & 34.1 & 29.5 \\
fashion & 33.8 & 35.0 & 43.3 & 40.9 \\
food\_processing & 29.6 & 31.6 & 38.1 & 36.6 \\
gas\_technology\_and\_engineering & 27.7 & 27.5 & 30.8 & 28.5 \\
geomatics & 34.9 & 36.6 & 41.6 & 38.9 \\
health & 40.0 & 44.0 & 59.0 & 52.0 \\
industrial\_engineer & 36.3 & 35.9 & 43.1 & 41.1 \\
information\_technology & 51.9 & 51.3 & 69.0 & 66.5 \\
interior\_architecture\_and\_design & 38.5 & 39.5 & 48.3 & 49.0 \\
korean\_history & 30.0 & 24.0 & 34.0 & 36.0 \\
law & 30.3 & 31.3 & 42.9 & 42.0 \\
machine\_design\_and\_manufacturing & 33.2 & 33.6 & 40.6 & 37.9 \\
management & 35.5 & 38.0 & 57.7 & 54.4 \\
maritime\_engineering & 36.7 & 39.2 & 44.2 & 43.0 \\
marketing & 57.4 & 57.7 & 74.6 & 74.9 \\
materials\_engineering & 30.2 & 30.1 & 39.4 & 36.9 \\
math & 26.7 & 29.0 & 29.7 & 31.0 \\
mechanical\_engineering & 29.9 & 28.6 & 35.5 & 30.0 \\
nondestructive\_testing & 33.0 & 34.2 & 42.6 & 39.0 \\
patent & 33.0 & 31.0 & 38.0 & 40.0 \\
political\_science\_and\_sociology & 36.0 & 37.0 & 55.0 & 51.7 \\
psychology & 28.3 & 29.9 & 44.1 & 41.4 \\
public\_safety & 30.8 & 29.4 & 34.1 & 30.2 \\
railway\_and\_automotive\_engineering & 33.0 & 32.0 & 33.7 & 29.4 \\
real\_estate & 37.0 & 37.5 & 44.5 & 44.0 \\
refrigerating\_machinery & 29.0 & 29.4 & 33.0 & 29.9 \\
social\_welfare & 37.0 & 37.2 & 55.1 & 53.9 \\
taxation & 30.5 & 33.5 & 42.5 & 44.0 \\
telecommunications\_and\_wireless\_technology & 41.9 & 41.5 & 55.3 & 51.7\\
\bottomrule
\end{tabular}
\label{tab:ap_yi}
\end{table}

\begin{table}[]
\fontsize{9}{11}\selectfont
\centering
\caption{5-shot accuracy using the Direct method for  \textsc{Qwen} (original and chat versions) broken down by category.}
\begin{tabular}{lcccccc}
\toprule
\textbf{Category} & \multicolumn{2}{c}{\textsc{Qwen-7B}} & \multicolumn{2}{c}{\textsc{Qwen-14B}} & \multicolumn{2}{c}{\textsc{Qwen-72B}} \\
 & Org. & Chat & Org. & Chat & Org. & Chat \\
 \midrule
accounting & 9.0 & 9.0 & 25.0 & 15.0 & 15.0 & 46.0 \\
agricultural\_sciences & 28.8 & 34.3 & 38.5 & 24.7 & 34.1 & 40.4 \\
aviation\_engineering\_and\_maintenance & 23.3 & 33.6 & 49.2 & 19.9 & 31.9 & 48.7 \\
biology & 15.2 & 29.0 & 40.5 & 15.4 & 26.5 & 39.7 \\
chemical\_engineering & 19.0 & 32.7 & 50.8 & 17.9 & 28.3 & 45.2 \\
chemistry & 24.3 & 44.2 & 54.3 & 21.2 & 37.7 & 50.7 \\
civil\_engineering & 17.6 & 31.3 & 46.5 & 17.5 & 31.7 & 46.7 \\
computer\_science & 32.0 & 54.2 & 75.7 & 30.6 & 52.8 & 76.4 \\
construction & 21.7 & 32.1 & 38.0 & 12.4 & 20.0 & 26.0 \\
criminal\_law & 4.5 & 12.5 & 40.0 & 4.0 & 9.0 & 36.5 \\
ecology & 34.2 & 46.2 & 52.4 & 35.3 & 45.7 & 53.1 \\
economics & 5.4 & 10.8 & 60.0 & 3.8 & 9.2 & 54.6 \\
education & 10.0 & 29.0 & 71.0 & 7.0 & 29.0 & 74.0 \\
electrical\_engineering & 22.3 & 27.7 & 34.8 & 18.8 & 26.9 & 35.0 \\
electronics\_engineering & 14.2 & 30.3 & 59.3 & 16.1 & 32.1 & 62.9 \\
energy\_management & 26.4 & 32.3 & 40.3 & 22.1 & 29.8 & 38.2 \\
environmental\_science & 26.4 & 32.6 & 38.0 & 28.0 & 34.4 & 41.4 \\
fashion & 32.6 & 42.8 & 49.6 & 29.6 & 41.6 & 48.7 \\
food\_processing & 8.3 & 19.0 & 45.8 & 5.7 & 12.9 & 36.8 \\
gas\_technology\_and\_engineering & 16.7 & 26.0 & 39.9 & 12.0 & 21.4 & 31.2 \\
geomatics & 22.8 & 31.8 & 43.8 & 19.5 & 29.4 & 41.8 \\
health & 9.0 & 30.0 & 71.0 & 13.0 & 28.0 & 61.0 \\
industrial\_engineer & 23.6 & 42.3 & 49.0 & 22.1 & 41.7 & 47.1 \\
information\_technology & 38.9 & 56.5 & 74.2 & 24.2 & 42.1 & 63.5 \\
interior\_architecture\_and\_design & 19.5 & 37.3 & 58.8 & 17.2 & 34.3 & 58.6 \\
korean\_history & 2.0 & 9.0 & 37.0 & 2.0 & 10.0 & 30.0 \\
law & 6.0 & 15.6 & 50.2 & 6.9 & 14.0 & 45.6 \\
machine\_design\_and\_manufacturing & 24.4 & 37.9 & 51.0 & 23.0 & 33.8 & 48.4 \\
management & 8.9 & 23.7 & 64.4 & 8.1 & 23.1 & 58.7 \\
maritime\_engineering & 21.3 & 40.8 & 49.8 & 18.0 & 31.8 & 43.3 \\
marketing & 37.8 & 59.7 & 85.1 & 37.1 & 60.2 & 81.9 \\
materials\_engineering & 15.1 & 29.0 & 50.2 & 7.2 & 20.9 & 37.9 \\
math & 18.7 & 26.7 & 36.7 & 20.3 & 22.7 & 28.7 \\
mechanical\_engineering & 12.8 & 26.1 & 41.5 & 14.5 & 25.4 & 46.4 \\
nondestructive\_testing & 27.2 & 40.9 & 48.4 & 26.4 & 38.7 & 48.5 \\
patent & 7.0 & 16.0 & 39.0 & 4.0 & 11.0 & 33.0 \\
political\_science\_and\_sociology & 11.7 & 30.7 & 62.0 & 13.3 & 27.3 & 56.7 \\
psychology & 18.4 & 31.1 & 51.5 & 15.2 & 30.1 & 45.4 \\
public\_safety & 7.9 & 14.1 & 40.3 & 7.5 & 16.0 & 41.0 \\
railway\_and\_automotive\_engineering & 20.5 & 31.7 & 40.1 & 22.2 & 31.6 & 39.2 \\
real\_estate & 2.0 & 7.5 & 53.0 & 3.5 & 8.5 & 45.0 \\
refrigerating\_machinery & 18.9 & 29.1 & 39.4 & 18.0 & 27.6 & 37.2 \\
social\_welfare & 25.0 & 41.0 & 64.7 & 22.0 & 38.2 & 60.1 \\
taxation & 3.0 & 7.5 & 42.5 & 4.0 & 7.5 & 32.0 \\
telecommunications\_and\_wireless\_technology & 39.1 & 50.0 & 64.2 & 36.6 & 50.7 & 64.4 \\ 
\bottomrule
\end{tabular}
\label{tab:ap_qw}
\end{table}

\begin{table}[ht!]
\fontsize{9}{11}\selectfont
\centering
\caption{5-shot accuracy using the Direct method for \textsc{Gemini-Pro}, \textsc{GPT-3.5-Turbo}, \textsc{GPT-4} and \textsc{HyperCLOVA X} broken down by category.}
\begin{tabular}{lcccc}
\toprule
\textbf{Category} & \textsc{Gemini-Pro} & \textsc{HCX.} & \textsc{GPT-3.5-Turbo} & \textsc{GPT-4} \\
\midrule
accounting & 44.0 & 46.0 & 42.0 & 71.0 \\
agricultural\_sciences & 42.5 & 42.7 & 34.1 & 50.2 \\
aviation\_engineering\_and\_maintenance & 53.0 & 49.0 & 43.5 & 63.9 \\
biology & 46.5 & 47.9 & 35.0 & 51.3 \\
chemical\_engineering & 51.9 & 47.9 & 42.9 & 61.3 \\
chemistry & 50.2 & 48.8 & 45.0 & 64.8 \\
civil\_engineering & 47.6 & 45.1 & 41.2 & 53.9 \\
computer\_science & 75.0 & 78.5 & 66.1 & 87.7 \\
construction & 37.6 & 41.9 & 34.9 & 46.7 \\
criminal\_law & 39.0 & 48.5 & 32.5 & 50.5 \\
ecology & 52.6 & 57.3 & 47.0 & 59.2 \\
economics & 53.1 & 65.4 & 40.8 & 67.7 \\
education & 58.0 & 72.0 & 40.0 & 84.0 \\
electrical\_engineering & 39.1 & 35.3 & 34.8 & 43.2 \\
electronics\_engineering & 60.2 & 59.8 & 52.1 & 69.9 \\
energy\_management & 38.1 & 37.6 & 33.9 & 43.9 \\
environmental\_science & 38.0 & 36.3 & 34.8 & 44.4 \\
fashion & 53.0 & 57.2 & 46.6 & 61.7 \\
food\_processing & 50.1 & 50.3 & 39.6 & 57.4 \\
gas\_technology\_and\_engineering & 42.0 & 42.3 & 34.5 & 49.0 \\
geomatics & 41.7 & 49.4 & 41.8 & 50.9 \\
health & 65.0 & 72.0 & 50.0 & 71.0 \\
industrial\_engineer & 50.7 & 50.2 & 43.3 & 58.1 \\
information\_technology & 72.3 & 73.1 & 66.3 & 83.7 \\
interior\_architecture\_and\_design & 63.5 & 69.1 & 51.0 & 69.8 \\
korean\_history & 41.0 & 42.0 & 32.0 & 35.0 \\
law & 48.5 & 58.7 & 40.2 & 58.6 \\
machine\_design\_and\_manufacturing & 54.4 & 50.8 & 43.9 & 64.9 \\
management & 59.7 & 64.3 & 51.2 & 74.1 \\
maritime\_engineering & 51.2 & 54.3 & 45.2 & 60.8 \\
marketing & 81.0 & 83.1 & 71.1 & 89.3 \\
materials\_engineering & 53.8 & 52.1 & 43.5 & 66.0 \\
math & 26.7 & 26.7 & 30.3 & 31.0 \\
mechanical\_engineering & 48.7 & 46.3 & 38.9 & 57.3 \\
nondestructive\_testing & 52.9 & 50.6 & 42.8 & 59.9 \\
patent & 37.0 & 52.0 & 34.0 & 43.0 \\
political\_science\_and\_sociology & 57.7 & 66.7 & 47.7 & 74.0 \\
psychology & 47.0 & 58.7 & 37.0 & 61.3 \\
public\_safety & 41.3 & 41.0 & 36.5 & 51.5 \\
railway\_and\_automotive\_engineering & 42.8 & 41.2 & 34.7 & 51.7 \\
real\_estate & 45.0 & 53.0 & 37.0 & 56.5 \\
refrigerating\_machinery & 40.7 & 40.0 & 33.9 & 48.1 \\
social\_welfare & 60.6 & 61.6 & 49.6 & 76.4 \\
taxation & 40.0 & 48.0 & 33.0 & 48.0 \\
telecommunications\_and\_wireless\_technology & 63.7 & 63.0 & 54.8 & 74.9 \\ 
\bottomrule
\end{tabular}

\label{tab:ap_pro}
\end{table}

\begin{table}[ht!]
\fontsize{9}{11}\selectfont
\centering
\caption{5-shot accuracy using the CoT method for \textsc{Qwen-72B-Chat}, \textsc{GPT-3.5-Turbo}, \textsc{GPT-4} and \textsc{HyperCLOVA X} broken down by category.}
\begin{tabular}{lcccc}
\toprule
\textbf{Category} & \textsc{Qwen-72B-Chat} & \textsc{HCX.} & \textsc{GPT-3.5-Turbo} & \textsc{GPT-4} \\
\midrule
accounting & 21.7 & 17.4 & 19.6 & 26.1 \\
agricultural\_sciences & 13.0 & 14.0 & 15.0 & 13.0 \\
aviation\_engineering\_and\_maintenance & 21.0 & 24.0 & 26.0 & 38.0 \\
biology & 21.0 & 24.0 & 15.0 & 14.0 \\
chemical\_engineering & 17.0 & 31.0 & 26.0 & 43.0 \\
chemistry & 22.0 & 30.0 & 29.0 & 44.0 \\
civil\_engineering & 17.0 & 25.0 & 20.0 & 16.0 \\
computer\_science & 25.0 & 36.0 & 18.0 & 25.0 \\
construction & 26.0 & 28.0 & 18.0 & 24.0 \\
criminal\_law & 9.0 & 24.0 & 9.0 & 8.0 \\
ecology & 12.0 & 24.0 & 16.0 & 11.0 \\
economics & 23.8 & 33.3 & 26.2 & 28.6 \\
education & 17.4 & 26.1 & 0.0 & 26.1 \\
electrical\_engineering & 11.0 & 24.0 & 20.0 & 30.0 \\
electronics\_engineering & 23.0 & 20.0 & 34.0 & 48.0 \\
energy\_management & 18.0 & 15.0 & 25.0 & 26.0 \\
environmental\_science & 16.0 & 22.0 & 17.0 & 27.0 \\
fashion & 20.0 & 29.0 & 24.0 & 16.0 \\
food\_processing & 17.0 & 24.0 & 21.0 & 28.0 \\
gas\_technology\_and\_engineering & 19.0 & 29.0 & 25.0 & 31.0 \\
geomatics & 18.0 & 24.0 & 20.0 & 24.0 \\
health & 8.7 & 26.1 & 26.1 & 21.7 \\
industrial\_engineer & 13.0 & 27.0 & 19.0 & 22.0 \\
information\_technology & 28.0 & 33.0 & 41.0 & 46.0 \\
interior\_architecture\_and\_design & 21.0 & 37.0 & 29.0 & 24.0 \\
korean\_history & 11.4 & 47.7 & 18.2 & 9.1 \\
law & 13.0 & 35.0 & 11.0 & 17.0 \\
machine\_design\_and\_manufacturing & 19.0 & 32.0 & 23.0 & 32.0 \\
management & 26.0 & 24.0 & 20.0 & 23.0 \\
maritime\_engineering & 21.0 & 27.0 & 19.0 & 21.0 \\
marketing & 29.0 & 18.0 & 17.0 & 18.0 \\
materials\_engineering & 21.0 & 24.0 & 20.0 & 24.0 \\
math & 18.0 & 32.0 & 31.0 & 51.0 \\
mechanical\_engineering & 17.0 & 25.0 & 20.0 & 36.0 \\
nondestructive\_testing & 19.0 & 23.0 & 27.0 & 24.0 \\
patent & 18.0 & 23.5 & 23.5 & 11.8 \\
political\_science\_and\_sociology & 24.4 & 27.8 & 4.4 & 14.4 \\
psychology & 16.0 & 36.0 & 14.0 & 9.0 \\
public\_safety & 21.0 & 30.0 & 13.0 & 12.0 \\
railway\_and\_automotive\_engineering & 12.0 & 25.0 & 19.0 & 29.0 \\
real\_estate & 10.1 & 25.8 & 10.1 & 14.6 \\
refrigerating\_machinery & 18.0 & 26.0 & 26.0 & 38.0 \\
social\_welfare & 13.0 & 35.0 & 36.0 & 51.0 \\
taxation & 5.2 & 26.0 & 10.4 & 4.2 \\
telecommunications\_and\_wireless\_technology & 25.0 & 30.0 & 30.0 & 38.0 \\ 
\bottomrule
\end{tabular}

\label{tab:ap_cot}
\end{table}

\end{document}